\documentclass{llncs}%
\pdfoutput=1
\usepackage{amsfonts}
\usepackage{amssymb}
\usepackage{graphicx}
\usepackage{amsmath}
\usepackage{color}
\usepackage{framed}%
\providecommand{\U}[1]{\protect\rule{.1in}{.1in}}
\definecolor{shadecolor}{gray}{0.9}
\begin{document}

\title{\textsc{Planning to Be Surprised: Optimal Bayesian Exploration in Dynamic
Environments}}
\author{Yi Sun, Faustino Gomez, and J\"{u}rgen Schmidhuber}
\institute{IDSIA}
\maketitle

\begin{abstract}
To maximize its success, an AGI typically needs to explore its initially
unknown world. Is there an optimal way of doing so? Here we derive an
affirmative answer for a broad class of environments.

\end{abstract}

\section{Introduction}

An intelligent agent is sent to explore an unknown environment. Over the
course of its mission, the agent makes observations, carries out actions, and
incrementally builds up a model of the environment from this interaction.
Since the way in which the agent selects actions may greatly affect the
efficiency of the exploration, the following question naturally arises:

\begin{quote}
\emph{How should the agent choose the actions such that the knowledge about
the environment accumulates as quickly as possible?}
\end{quote}

In this paper, this question is addressed under a classical framework, in
which the agent improves its model of the environment through probabilistic
inference, and learning progress is measured in terms of Shannon information
gain. We show that the agent can, at least in principle, optimally choose
actions based on previous experiences, such that the cumulative expected
information gain is maximized. We then consider a special case, namely
exploration in finite MDPs, where we demonstrate, both in theory and through
experiment, that the optimal Bayesian exploration strategy can be effectively
approximated by solving a sequence of dynamic programming problems.

The rest of the paper is organized as follows: Section 2 reviews the basic
concepts and establishes the terminology; Section 3 elaborates the principle
of optimal Bayesian exploration; Section 4 focuses on exploration in finite
MDP; Section 5 presents a simple experiment; The related works are briefly
reviewed in Section 6; Section 7 concludes the paper.

\section{Preliminaries}

Suppose that the agent interacts with the environment in discrete time cycles
$t=1,2,\ldots$. In each cycle, the agent performs an action $a$, then receives
a sensory input $o$. A \emph{history} $h$ is either the empty string
$\emptyset$ or a string of the form $a_{1}o_{1}\cdots a_{t}o_{t}$ for some
$t$, and $ha$ and $hao$ refer to the strings resulting from appending $a$ and
$ao$ to $h$, respectively.

\subsection{Learning from Sequential Interactions}

To facilitate the subsequent discussion under a probabilistic framework, we
make the following assumptions:

\begin{description}
\item[Assumption I.] The models of the environment under consideration are
fully described by a random element $\Theta$ which \emph{depends solely on the
environment}. Moreover, the agent's initial knowledge about $\Theta$ is
summarized by a prior density $p\left(  \theta\right)  $.

\item[Assumption II.] The agent is equipped with a \emph{conditional
predictor} $p\left(  o|ha;\theta\right)  $, i.e.\ the agent is capable of
refining its prediction in the light of information about $\Theta$.
\end{description}

Using $p\left(  \theta\right)  $ and $p\left(  o|ha;\theta\right)  $ as
building blocks, it is straightforward to formulate learning in terms of
probabilistic inference. From Assumption \textbf{I}, given the history $h$,
the agent's knowledge about $\Theta$ is fully summarized by $p\left(
\theta|h\right)  $. According to Bayes rule, $p\left(  \theta|hao\right)
=\frac{p\left(  \theta|ha\right)  p\left(  o|ha;\theta\right)  }{p\left(
o|ha\right)  }$, with $p\left(  o|ha\right)  =\int p\left(  o|ha,\theta
\right)  p\left(  \theta|h\right)  d\theta$. The term $p\left(  \theta
|ha\right)  $ represents the agent's current knowledge about $\Theta$ given
history $h$ and an additional action $a$. Since $\Theta$ depends solely on the
environment, and, importantly, \emph{knowing the action without subsequent
observations cannot change the agent's state of knowledge about }$\Theta$,
$p\left(  \theta|ha\right)  =p\left(  \theta|h\right)  $, hence the knowledge
about $\Theta$ can be updated using%
\begin{equation}
p\left(  \theta|hao\right)  =p\left(  \theta|h\right)  \cdot\frac{p\left(
o|ha;\theta\right)  }{p\left(  o|ha\right)  }\text{.} \label{eq-01}%
\end{equation}

It is worth pointing out that $p\left(  o|ha;\theta\right)  $ is chosen
\textit{a priori}. It is not required that they match the true dynamics of the
environment, but the effectiveness of the learning certainly depends on the
choices of $p\left(  o|ha;\theta\right)  $. For example, if $\Theta
\in\mathbb{R}$, and $p\left(  o|ha;\theta\right)  $ depends on $\theta$ only
through its sign, then no knowledge other than the sign of $\Theta$ can be learned.

\subsection{Information Gain as Learning Progress}

Let $h$ and $h^{\prime}$ be two histories such that $h$ is a prefix of
$h^{\prime}$. The respective posterior of $\Theta$ are $p\left(
\theta|h\right)  $ and $p\left(  \theta|h^{\prime}\right)  $. Using $h$ as a
reference point, the amount of information gained when the history grows to
$h^{\prime}$, can be measured using the KL divergence between $p\left(
\theta|h\right)  $ and $p\left(  \theta|h^{\prime}\right)  $. This
\emph{information gain} from $h$ to $h^{\prime}$ is defined as%
\[
g(h^{\prime}\Vert h)=KL\left(  p\left(  \theta|h^{\prime}\right)  \Vert
p\left(  \theta|h\right)  \right)  =\int p\left(  \theta|h^{\prime}\right)
\log\frac{p\left(  \theta|h^{\prime}\right)  }{p\left(  \theta|h\right)
}d\theta\text{.}%
\]
As a special case, if $h=\emptyset$, then $g\left(  h^{\prime}\right)
=g\left(  h^{\prime}\Vert\emptyset\right)  $ is the \emph{cumulative
information gain} with respect to the prior $p\left(  \theta\right)  $. We
also write $g\left(  ao\Vert h\right)  $ for $g\left(  hao\Vert h\right)  $,
which denotes the information gained from an additional action-observation pair.

From an information theoretic point of view, the KL divergence between two
distributions $p$ and $q$ represents the additional number of bits required to
encode elements sampled from $p$, using optimal coding strategy designed for
$q$. This can be interpreted as the degree of `unexpectedness' or `surprise'
caused by observing samples from $p$ when expecting samples from $q$.

The key property information gain for the treatment below is the following
decomposition: Let $h$ be a prefix of $h^{\prime}$ and $h^{\prime}$ be a
prefix of $h^{\prime\prime}$, then%
\begin{align*}
&  \mathbb{E}_{h^{\prime\prime}|h^{\prime}}g\left(  h^{\prime\prime}\Vert
h\right) \\
&  =\mathbb{E}_{h^{\prime\prime}|h^{\prime}}\int p\left(  \theta
|h^{\prime\prime}\right)  \log\frac{p\left(  \theta|h^{\prime\prime}\right)
}{p\left(  \theta|h\right)  }d\theta\\
&  =\mathbb{E}_{h^{\prime\prime}|h^{\prime}}\int p\left(  \theta
|h^{\prime\prime}\right)  \left[  \log\frac{p\left(  \theta|h^{\prime\prime
}\right)  }{p\left(  \theta|h^{\prime}\right)  }+\log\frac{p\left(
\theta|h^{\prime}\right)  }{p\left(  \theta|h\right)  }\right]  d\theta\\
&  =\mathbb{E}_{h^{\prime\prime}|h^{\prime}}\int p\left(  \theta
|h^{\prime\prime}\right)  \log\frac{p\left(  \theta|h^{\prime\prime}\right)
}{p\left(  \theta|h^{\prime}\right)  }d\theta+\int\mathbb{E}_{h^{\prime\prime
}|h^{\prime}}p\left(  \theta|h^{\prime\prime}\right)  \log\frac{p\left(
\theta|h^{\prime}\right)  }{p\left(  \theta|h\right)  }d\theta\\
&  =\mathbb{E}_{h^{\prime\prime}|h^{\prime}}g\left(  h^{\prime\prime}\Vert
h^{\prime}\right)  +\int\mathbb{E}_{h^{\prime\prime}|h^{\prime}}p\left(
\theta|h^{\prime\prime}\right)  \log\frac{p\left(  \theta|h^{\prime}\right)
}{p\left(  \theta|h\right)  }d\theta\text{.}%
\end{align*}
From updating formula Eq.\ref{eq-01},%
\begin{align*}
\sum_{o}p\left(  o|ha\right)  p\left(  \theta|hao\right)   &  =\sum
_{o}p\left(  \theta|h\right)  p\left(  o|ha,\theta\right) \\
&  =p\left(  \theta|h\right)  \sum_{o}p\left(  o|ha,\theta\right) \\
&  =p\left(  \theta|h\right)  \text{.}%
\end{align*}
Using this relation recursively,%
\begin{align*}
\mathbb{E}_{h^{\prime\prime}|h^{\prime}}p\left(  \theta|h^{\prime\prime
}\right)   &  =\sum_{a_{1}}\sum_{o_{1}}\cdots\sum_{a_{t}}\sum_{o_{t}}p\left(
\theta|h^{\prime}a_{1}o_{1}\cdots a_{t}o_{t}\right) \\
&  =\sum_{a_{1}}\sum_{o_{1}}\cdots\sum_{a_{t-1}}\sum_{o_{t-1}}p\left(
\theta|h^{\prime}a_{1}o_{1}\cdots a_{t-1}o_{t-1}\right) \\
&  =\cdots\\
&  =p\left(  \theta|h^{\prime}\right)  \text{,}%
\end{align*}
therefore%
\begin{equation}
\mathbb{E}_{h^{\prime\prime}|h^{\prime}}g\left(  h^{\prime\prime}\Vert
h\right)  =g\left(  h^{\prime}\Vert h\right)  +\mathbb{E}_{h^{\prime\prime
}|h^{\prime}}g\left(  h^{\prime\prime}\Vert h^{\prime}\right)  \text{.}
\label{eq-02}%
\end{equation}
That is, the information gain is \emph{additive in expectation}.

Having defined the information gain from trajectories ending with
observations, one may proceed to define the \emph{expected} \emph{information
gain} of performing action $a$, before observing the outcome $o$. Formally,
the \emph{expected information gain} of performing $a$ with respect to the
current history $h$ is given by $g\left(  a\Vert h\right)  =\mathbb{E}%
_{o|ha}g\left(  ao\Vert h\right)  $. A simple derivation gives%
\begin{align*}
g\left(  a\Vert h\right)   &  =\sum_{o}p\left(  o|ha\right)  \int p\left(
\theta|hao\right)  \log\frac{p\left(  \theta|hao\right)  }{p\left(
\theta|h\right)  }d\theta\\
&  =\sum_{o}\int p\left(  o,\theta|ha\right)  \log\frac{p\left(
\theta|hao\right)  p\left(  o|ha\right)  }{p\left(  \theta|h\right)  p\left(
o|ha\right)  }d\theta\\
&  =\sum_{o}\int p\left(  o,\theta|ha\right)  \log\frac{p\left(
o,\theta|ha\right)  }{p\left(  \theta|ha\right)  p\left(  o|ha\right)
}d\theta\\
&  =I\left(  O;\Theta|ha\right)  \text{,}%
\end{align*}
which means that $g\left(  a\Vert h\right)  $ is the mutual information
between $\Theta$ and the random variable $O$ representing the unknown
observation, conditioned on the history $h$ and action $a$.\footnote{Side
note: To generalize the discussion, concepts from algorithmic information
theory, such as compression distance, may also be used here. However,
restricting the discussion under a probabilistic framework greatly simplifies
the matter.}

\section{Optimal Bayesian Exploration}

In this section, the general principle of optimal Bayesian exploration in
dynamic environments is presented. We first give results obtained by assuming
a fixed limited life span for our agent, then discuss a condition required to
extend this to infinite time horizons.

\subsection{Results for Finite Time Horizon}

Suppose that the agent has experienced history $h$, and is about to choose
$\tau$ more actions in the future. Let $\pi$ be a policy mapping the set of
histories to the set of actions, such that the agent performs $a$ with
probability $\pi\left(  a|h\right)  $ given $h$. Define the \emph{curiosity
Q-value} $q_{\pi}^{\tau}\left(  h,a\right)  $ as the expected information
gained from the additional $\tau$ actions, assuming that the agent performs
$a$ in the next step and follows policy $\pi$ in the remaining $\tau-1$ steps.
Formally, for $\tau=1$,%
\[
q_{\pi}^{1}\left(  h,a\right)  =\mathbb{E}_{o|ha}g\left(  ao\Vert h\right)
=g\left(  a\Vert h\right)  \text{,}%
\]
and for $\tau>1$,%
\begin{align*}
q_{\pi}^{\tau}\left(  h,a\right)   &  =\mathbb{E}_{o|ha}\mathbb{E}_{a_{1}%
|hao}\mathbb{E}_{o_{1}|haoa_{1}}\cdots\mathbb{E}_{o_{\tau-1}|h\cdots
a_{\tau-1}}g\left(  haoa_{1}o_{1}\cdots a_{\tau-1}o_{\tau-1}\Vert h\right)  \\
&  =\mathbb{E}_{o|ha}\mathbb{E}_{a_{1}o_{1}\cdots a_{\tau-1}o_{\tau-1}%
|hao}g\left(  haoa_{1}o_{1}\cdots a_{\tau-1}o_{\tau-1}\Vert h\right)  \text{.}%
\end{align*}
\newline

The curiosity Q-value can be defined recursively. Applying Eq.~\ref{eq-02} for
$\tau=2$,%
\begin{align*}
q_{\pi}^{\tau}\left(  h,a\right)   &  =\mathbb{E}_{o|ha}\mathbb{E}_{a_{1}%
o_{1}|hao}g\left(  haoa_{1}o_{1}\Vert h\right)  \\
&  =\mathbb{E}_{o|ha}\left[  g\left(  ao\Vert h\right)  +\mathbb{E}%
_{a_{1}o_{1}|hao}g\left(  a_{1}o_{1}\Vert hao\right)  \right]  \\
&  =g\left(  a\Vert h\right)  +\mathbb{E}_{o|ha}\mathbb{E}_{a^{\prime}%
|hao}q_{\pi}^{1}\left(  hao,a^{\prime}\right)  \text{.}%
\end{align*}
And for $\tau>2$,%
\begin{align}
q_{\pi}^{\tau}\left(  h,a\right)   &  =\mathbb{E}_{o|ha}\mathbb{E}_{a_{1}%
o_{1}\cdots a_{\tau-1}o_{\tau-1}|hao}g\left(  haoa_{1}o_{1}\cdots a_{\tau
-1}o_{\tau-1}\Vert h\right)  \nonumber\\
&  =\mathbb{E}_{o|ha}\left[  g\left(  ao\Vert h\right)  +\mathbb{E}%
_{a_{1}o_{1}\cdots a_{\tau-1}o_{\tau-1}}g\left(  haoa_{1}o_{1}\cdots
a_{\tau-1}o_{\tau-1}\Vert hao\right)  \right]  \nonumber\\
&  =g\left(  a\Vert h\right)  +\mathbb{E}_{o|ha}\mathbb{E}_{a^{\prime}%
|hao}q_{\pi}^{\tau-1}\left(  hao,a^{\prime}\right)  \text{.}\label{eq-03}%
\end{align}
Noting that Eq.\ref{eq-03} bears great resemblance to the definition of
state-action values ($Q(s,a)$) in reinforcement learning, one can similarly
define the \emph{curiosity value} of a particular history as $v_{\pi}^{\tau
}\left(  h\right)  =\mathbb{E}_{a|h}q_{\pi}^{\tau}\left(  h,a\right)  $,
analogous to state values ($V(s)$), which can also be iteratively defined as
$v_{\pi}^{1}\left(  h\right)  =\mathbb{E}_{a|h}g\left(  a\Vert h\right)  $,
and%
\[
v_{\pi}^{\tau}\left(  h\right)  =\mathbb{E}_{a|h}\left[  g\left(  a\Vert
h\right)  +\mathbb{E}_{o|ha}v_{\pi}^{\tau-1}\left(  hao\right)  \right]
\text{.}%
\]
The curiosity value $v_{\pi}^{\tau}\left(  h\right)  $ is the expected
information gain of performing the additional $\tau$ steps, assuming that the
agent follows policy $\pi$. The two notations can be combined to write%
\begin{equation}
q_{\pi}^{\tau}\left(  h,a\right)  =g\left(  a\Vert h\right)  +\mathbb{E}%
_{o|ha}v_{\pi}^{\tau-1}\left(  hao\right)  \text{.}\label{eq-04}%
\end{equation}
This equation has an interesting interpretation: since the agent is operating
in a dynamic environment, it has to take into account not only the immediate
expected information gain of performing the current action, i.e., $g\left(
a\Vert h\right)  $, but also the expected curiosity value of the situation in
which the agent ends up due to the action, i.e., $v_{\pi}^{\tau-1}\left(
hao\right)  $. As a consequence,\emph{ the agent needs to choose actions that
balance the two factors in order to improve its total expected information
gain}.

Now we show that there is a optimal policy $\pi_{\ast}$, which leads to the
maximum cumulative expected information gain given any history $h$. To obtain
the optimal policy, one may work backwards in $\tau$, taking greedy actions
with respect to the curiosity Q-values at each time step. Namely, for $\tau
=1$, let%
\[
q^{1}\left(  h,a\right)  =g\left(  a\Vert h\right)  \text{, }\pi_{\ast}%
^{1}\left(  h\right)  =\arg\max_{a}g\left(  a\Vert h\right)  \text{, and
}v^{1}\left(  h\right)  =\max_{a}g\left(  a\Vert h\right)  \text{,}%
\]
such that $v^{1}\left(  h\right)  =q^{1}\left(  h,\pi^{1}\left(  h\right)
\right)  $, and for $\tau>1$, let%
\[
q^{\tau}\left(  h,a\right)  =g\left(  a\Vert h\right)  +\mathbb{E}%
_{o|ha}\left[  \max_{a^{\prime}}q^{\tau-1}\left(  a^{\prime}|hao\right)
\right]  =g\left(  a\Vert h\right)  +\mathbb{E}_{o|ha}v^{\tau-1}\left(
hao\right)  \text{,}%
\]
with $\pi_{\ast}^{\tau}\left(  h\right)  =\arg\max_{a}q^{\tau}\left(
h,a\right)  $ and $v^{\tau}\left(  h\right)  =\max_{a}q^{\tau}\left(
h,a\right)  $. We show that $\pi_{\ast}^{\tau}\left(  h\right)  $ is indeed
the optimal policy for any given $\tau$ and $h$ in the sense that the
curiosity value, when following $\pi_{\ast}^{\tau}$, is maximized. To see
this, take any other strategy $\pi$, first notice that%
\[
v^{1}\left(  h\right)  =\max_{a}g\left(  a\Vert h\right)  \geq\mathbb{E}%
_{a|h}g\left(  a\Vert h\right)  =v_{\pi}^{1}\left(  h\right)  \text{.}%
\]
Moreover, assuming $v^{\tau}\left(  h\right)  \geq v_{\pi}^{\tau}\left(
h\right)  $,%
\begin{align*}
v^{\tau+1}\left(  h\right)   &  =\max_{a}\left[  g\left(  a\Vert h\right)
+\mathbb{E}_{o|ha}v^{\tau}\left(  hao\right)  \right]  \geq\max_{a}\left[
g\left(  a\Vert h\right)  +\mathbb{E}_{o|ha}v_{\pi}^{\tau}\left(  hao\right)
\right]  \\
&  \geq\mathbb{E}_{a|h}\left[  g\left(  a\Vert h\right)  +\mathbb{E}%
_{o|ha}v_{\pi}^{\tau}\left(  hao\right)  \right]  =v_{\pi}^{\tau+1}\left(
h\right)  \text{.}%
\end{align*}
Therefore $v^{\tau}\left(  h\right)  \geq v_{\pi}^{\tau}\left(  h\right)  $
holds for arbitrary $\tau$, $h$, and $\pi$. The same can be shown for
curiosity Q-values, namely, $q^{\tau}\left(  h,a\right)  \geq q_{\pi}^{\tau
}\left(  h,a\right)  $, for all $\tau$, $h$, $a$, and $\pi$. It may be
beneficial to write $q^{\tau}$ in explicit forms, namely,%
\[
q^{\tau}\left(  h,a\right)  =\mathbb{E}_{o|ha}\max_{a_{1}}\mathbb{E}%
_{o_{1}|haoa_{1}}\cdots\max_{a_{\tau-1}}\mathbb{E}_{o_{\tau-1}|h\cdots
a_{\tau-1}}g\left(  haoa_{1}o_{1}\cdots a_{\tau-1}o_{\tau-1}\Vert h\right)
\text{.}%
\]

Now consider that the agent has a fixed life span $T$. It can be seen that at
time $t$, the agent has to perform $\pi_{\ast}^{T-t}\left(  h_{t-1}\right)  $
to maximize the expected information gain in the remaining $T-t$ steps. Here
$h_{t-1}=a_{1}o_{1}\cdots a_{t-1}o_{t-1}$ is the history at time $t$. However,
from Eq.\ref{eq-02},%
\[
\mathbb{E}_{h_{T}|h_{t-1}}g\left(  h_{T}\right)  =g\left(  h_{t-1}\right)
+\mathbb{E}_{h_{T}|h_{t-1}}g\left(  h_{T}\Vert h_{t-1}\right)  \text{.}%
\]
Note that at time $t$, $g\left(  h_{t-1}\right)  $ is a constant, thus
\emph{maximizing the cumulative expected information gain in the remaining
time steps is equivalent to maximizing the expected information gain of the
whole trajectory with respect to the prior}. The result is summarized in the
following proposition:

\begin{proposition}
\label{prop-00}Let $q_{1}\left(  h,a\right)  =\bar{g}\left(  a\Vert h\right)
$, $v_{1}\left(  h\right)  =\max_{a}q_{1}\left(  h,a\right)  $, and%
\[
q_{\tau}\left(  h,a\right)  =\bar{g}\left(  a\Vert h\right)  +\mathbb{E}%
_{o|ha}v_{\tau-1}\left(  hao\right)  \text{, }v_{\tau}\left(  h\right)
=\max_{a}q_{\tau}\left(  h,a\right)  \text{,}%
\]
then the policy $\pi_{\tau}^{\ast}\left(  h\right)  =\arg\max_{a}q_{\tau
}\left(  h,a\right)  $ is optimal in the sense that $v_{\tau}\left(  h\right)
\geq v_{\tau}^{\pi}\left(  h\right)  $, $q_{\tau}\left(  h,a\right)  \geq
q_{\tau}^{\pi}\left(  h,a\right)  $ for any $\pi$, $\tau$, $h$ and $a$.
In particular, for an agent with fixed life span $T$, following $\pi
_{T-t}^{\ast}\left(  h_{t-1}\right)  $ at time $t=1,\ldots,T$ is optimal in
the sense that the expected cumulative information gain with respect to the
prior is maximized.
\end{proposition}

\subsection{Non-triviality of the Result}

Intuitively, the interpretation of the recursive definition of the curiosity
(Q) value is simple, and bears clear resemblance to their counterparts in
reinforcement learning. It might be tempting to think that the result is
nothing more than solving the finite horizon reinforcement learning problem
using $g\left(  a\Vert h\right)  $ or $g\left(  ao\Vert h\right)  $ as the
reward signals. However, this is not the case.

First, note that the decomposition Eq.\ref{eq-02} is a direct consequence of
the formulation of the KL divergence. The decomposition does not necessarily
hold if $g\left(  h\right)  $ is replaced with other types of measures of
information gain.

Second, it is worth pointing out that $g\left(  ao\Vert h\right)  $ and
$g\left(  a\Vert h\right)  $ behave differently from normal reward signals in
the sense that they are \emph{additive only in expectation}, while in the
reinforcement learning setup, the reward signals are usually assumed to be
additive, i.e., adding reward signals together is always meaningful. Consider
a simple problem with only two actions. If $g\left(  ao\Vert h\right)  $ is a
plain reward function, then $g\left(  ao\Vert h\right)  +g\left(  a^{\prime
}o^{\prime}\Vert hao\right)  $ should be meaningful, no matter if $a$ and $o$
is known or not. But this is not the case, since the sum does not have a valid
information theoretic interpretation. On the other hand, the sum is meaningful
\emph{in expectation}. Namely, when $o$ has \emph{not} been observed, from
Eq.\ref{eq-02},%
\[
g\left(  ao\Vert h\right)  +\mathbb{E}_{o^{\prime}|haoa^{\prime}}g\left(
a^{\prime}o^{\prime}\Vert hao\right)  =\mathbb{E}_{o^{\prime}|haoa^{\prime}%
}g\left(  aoa^{\prime}o^{\prime}\Vert h\right)  \text{,}%
\]
the sum can be interpreted as the expectation of the information gained from
$h$ to $haoa^{\prime}o^{\prime}$. This result shows that $g\left(  a\Vert
h\right)  $ or $g\left(  ao\Vert h\right)  $ can be treated as additive reward
signals only when one is planning ahead.

\begin{figure}[tb]
\begin{center}
\includegraphics[
width=3.5in
]{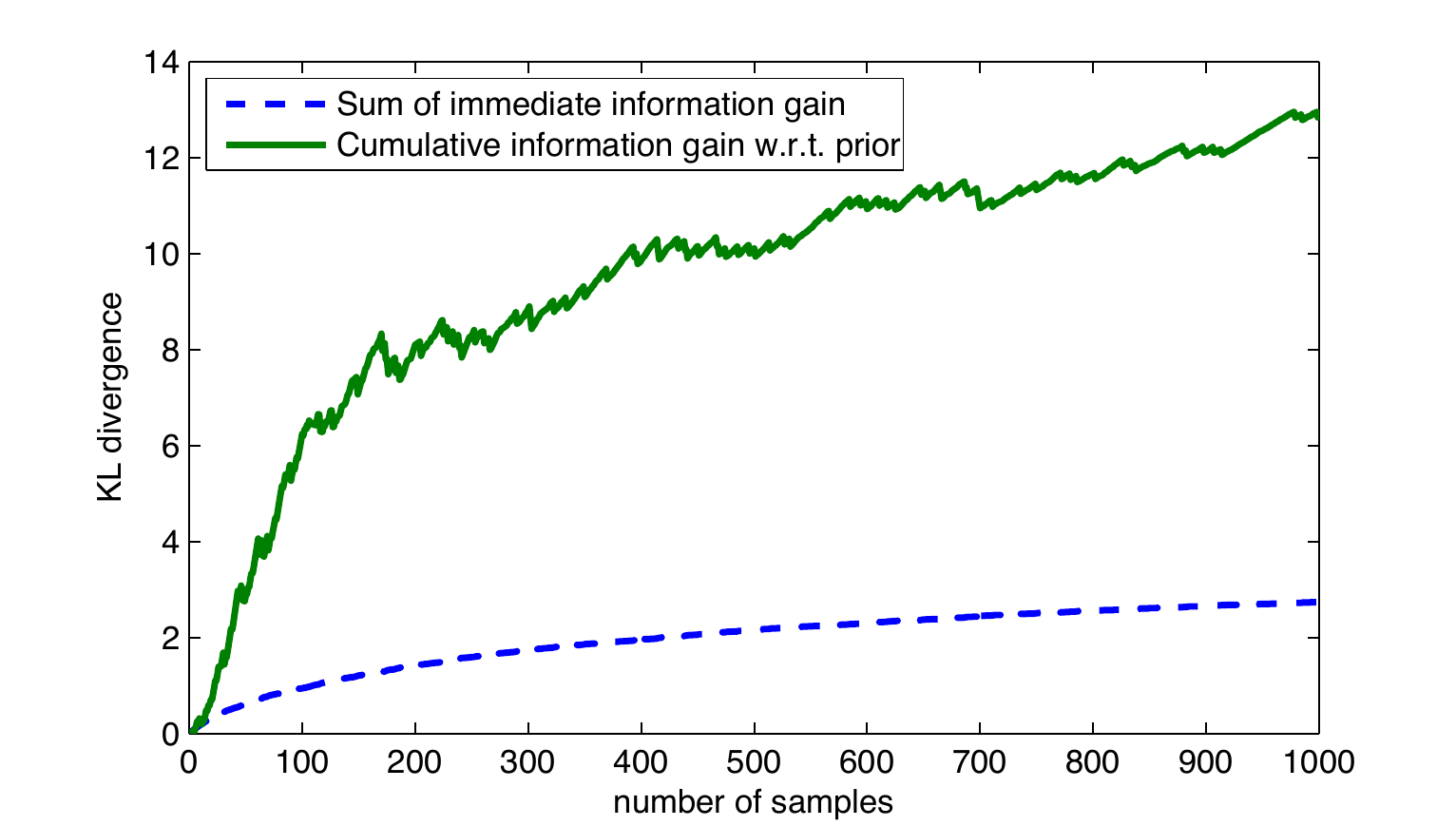}\vspace*{-2mm}
\end{center}
\caption{Illustration of the difference between the sum of one-step
information gain and the cumulative information gain with respect to the
prior. In this case, $1000$ independent samples are generated from a
distribution over finite sample space $\left\{  1,2,3\right\}  $, with
$p\left(  x=1\right)  =0.1$, $p\left(  x=2\right)  =0.5$, and $p\left(
x=3\right)  =0.4$. The task of learning is to recover the mass function from
the samples, assuming a Dirichlet prior $Dir\left(  \frac{50}{3},\frac{50}%
{3},\frac{50}{3}\right)  $. The KL divergence between two Dirichlet
distributions are computed according to \cite{01-KLDir}. It is clear from the
graph that the cumulative information gain fluctuates when the number of
samples increases, while the sum of the one-step information gain increases
monotonically. It also shows that the difference between the two quantities
can be large.}%
\label{fig-1}%
\end{figure}

To emphasize the difference further, note that all immediate information gains
$g\left(  ao\Vert h\right)  $ are non-negative since they are essentially KL
divergence. A natural assumption would be that the information gain $g\left(
h\right)  $, which is the sum of all $g\left(  ao\Vert h\right)  $ in
expectation, grows monotonically when the length of the history increases.
However, this is not the case, see Figure \ref{fig-1} for example. Although
$g\left(  ao\Vert h\right)  $ is always non-negative, some of the gain may
pull $\theta$ closer to its prior density $p\left(  \theta\right)  $,
resulting in a decrease of KL divergence between $p\left(  \theta|h\right)  $
and $p\left(  \theta\right)  $. This is never the case if one considers the
normal reward signals in reinforcement learning, where the accumulated reward
would never decrease if all rewards are non-negative.

\subsection{The Algorithm}

The definition of the optimal exploration policy is constructive, which means
that it can be readily implemented, provided that the number of actions and
possible observations is finite so that the expectation and maximization can
be computed exactly.

The following two algorithms computes the maximum curiosity value $v^{\tau
}\left(  h\right)  $ and the maximum curiosity Q-value $q^{\tau}\left(
h,a\right)  $, respectively, assuming that the expected immediate gain
$g\left(  a\Vert h\right)  $ can be computed.

\begin{center}%
\[%
\begin{tabular}
[c]{r|l}\hline
\multicolumn{2}{l}{\texttt{CuriosityValue}$(h,\tau)$}\\
\multicolumn{2}{l}{Input: history $h$, look-ahead $\tau$}\\
\multicolumn{2}{l}{Output: curiosity value $v^{\tau}\left(  h\right)  $%
}\\\hline\hline
1 & $v\leftarrow0$\\
2 & \textbf{for} \textit{all possible} $a$\\
3 & $\ \ v\leftarrow$ $\max\left(  v,\text{\texttt{CuriosityQValue}}\left(
h,a,\tau\right)  \right)  $\\
4 & \textbf{end} \textbf{for}\\
5 & \textbf{return} $v$\\\hline
\end{tabular}
\ \
\]

\[%
\begin{tabular}
[c]{r|l}\hline
\multicolumn{2}{l}{\texttt{CuriosityQValue}$(h,a,\tau)$}\\
\multicolumn{2}{l}{Input: history $h$, action $a$, look-ahead $\tau$}\\
\multicolumn{2}{l}{Output: curiosity Q-value $q^{\tau}\left(  h,a\right)  $%
}\\\hline\hline
1 & $q\leftarrow g\left(  a\Vert h\right)  $\\
2 & \textbf{if} $\tau\neq0$\\
3 & \ \ \textbf{for} \textit{all possible} $o$\\
4 & $\ \ \ \ q\leftarrow q+p\left(  o|ha\right)  \cdot$\texttt{CuriosityValue}%
$\left(  hao,\tau-1\right)  $\\
5 & \ \ \textbf{end} \textbf{for}\\
6 & \textbf{end} \textbf{if}\\
7 & \textbf{return} $q$\\\hline
\end{tabular}
\ \
\]

\end{center}

The complexity of both \texttt{CuriosityValue} and \texttt{CuriosityQValue}
are $O\left(  \left(  n_{o}n_{a}\right)  ^{\tau}\right)  $, where $n_{o}$ and
$n_{a}$ are the number of possible observations and actions, respectively.
Since the cost is exponential on $\tau$, planning with large number of look
ahead steps is infeasible, and approximation heuristics must be used in practice.

\subsection{Extending to Infinite Horizon}

Having to restrict the maximum life span of the agent is rather inconvenient.
It is tempting to define the curiosity Q-value in the infinite time horizon
case as the limit of curiosity Q-values with increasing life spans,
$T\rightarrow\infty$. However, this cannot be achieved without additional
technical constraints. For example, consider simple coin tossing. Assuming a
$Beta\left(  1,1\right)  $ over the probability of seeing heads, then the
expected cumulative information gain for the next $T$ flips is given by%
\[
v^{T}\left(  h_{1}\right)  =I\left(  \Theta;X_{1},\ldots,X_{T}\right)
\sim\log T\text{.}%
\]
With increasing $T$, $v^{T}\left(  h_{1}\right)  \rightarrow\infty$. A
frequently used approach to simplifying the math is to introduce a discount
factor $\gamma$, as used in reinforcement learning. Assume that the agent has
a maximum $\tau$ actions left, but before finishing the $\tau$ actions it may
be forced to leave the environment with probability $1-\gamma$ ($0\leq
\gamma<1$) at each time step. In this case, the curiosity Q-value becomes
$q_{\pi}^{1,\gamma}\left(  h,a\right)  =g\left(  a\Vert h\right)  $, and%
\begin{align*}
q_{\pi}^{\tau,\gamma}\left(  h,a\right)   &  =\left(  1-\gamma\right)
g\left(  a\Vert h\right)  +\gamma\left[  g\left(  a\Vert h\right)
+\mathbb{E}_{o|ha}\mathbb{E}_{a^{\prime}|hao}q_{\pi}^{\tau-1,\gamma}\left(
hao,a^{\prime}\right)  \right]  \\
&  =g\left(  a\Vert h\right)  +\gamma\mathbb{E}_{o|ha}\mathbb{E}_{a^{\prime
}|hao}q_{\pi}^{\tau-1,\gamma}\left(  hao,a^{\prime}\right)  \text{.}%
\end{align*}
One may also interpret $q_{\pi}^{\tau,\gamma}\left(  h,a\right)  $ as a linear
combination of curiosity Q-values without the discount,
\[
q_{\pi}^{\tau,\gamma}\left(  h,a\right)  =\left(  1-\gamma\right)  \sum
_{t=1}^{\tau}\gamma^{t-1}q_{\pi}^{t}\left(  h,a\right)  +\gamma^{\tau}q_{\pi
}^{\tau}\left(  h,a\right)  \text{,}%
\]
Note that curiosity Q-values with larger look-ahead steps are weighed
exponentially less.

The optimal policy in the discounted case is given by%
\[
q^{1,\gamma}\left(  h,a\right)  =g\left(  a\Vert h\right)  \text{,
}v^{1,\gamma}\left(  h\right)  =\max_{a}q^{1,\gamma}\left(  h,a\right)
\text{,}%
\]
and%
\[
q^{\tau,\gamma}\left(  h,a\right)  =g\left(  a\Vert h\right)  +\gamma
\mathbb{E}_{o|ha}v^{\tau-1,\gamma}\left(  hao\right)  \text{, }v^{\tau,\gamma
}\left(  h\right)  =\max_{a}q^{\tau,\gamma}\left(  h,a\right)  \text{.}%
\]
The optimal actions are given by $\pi_{\ast}^{\tau,\gamma}\left(  h\right)
=\arg\max_{a}q^{\tau,\gamma}\left(  h,a\right)  $. The proof that $\pi_{\ast
}^{\tau,\gamma}$ is optimal is similar to the one for no-discount case and
thus is omitted here.

Adding the discount enables one to define the curiosity Q-value in infinite
time horizon in a number of cases. However, it is still possible to construct
scenarios where such discount fails. Consider a infinite list of bandits. For
bandit $n$, there are $n$ possible outcomes with Dirichlet prior $Dir\left(
\frac{1}{n},\ldots,\frac{1}{n}\right)  $. The expected information gain of
pulling bandit $n$ for the first time is then given by%
\[
\log n-\psi\left(  2\right)  +\log\left(  1+\frac{1}{n}\right)  \sim\log
n\text{.}%
\]
Assume at time $t$, only the first $e^{e^{2t}}$ bandits are available, thus
the curiosity Q-value in finite time horizon is always finite. However, since
the largest expected information gain grows at speed $e^{t^{2}}$, for any
given $\gamma>0$, $q^{\tau,\gamma}$ goes to infinity with increasing $\tau$.
This example gives the intuition that to make the curiosity Q-value
meaningful, the `total information content' of the environment (or its growing
speed) must be bounded.

The following two Lemmas are useful for later discussion.

\begin{lemma}
\label{lem-01}$q_{\pi}^{\tau+1,\gamma}\left(  h,a\right)  -q_{\pi}%
^{\tau,\gamma}\left(  h,a\right)  =\gamma^{\tau}\mathbb{E}_{oa_{1}\cdots
o_{\tau-1}a_{\tau}|ha}g\left(  a_{\tau}\Vert h\cdots o_{\tau-1}\right)  $.
\end{lemma}

%

%TCIMACRO{\TeXButton{beginshaded}{\begin{shaded}}}%
%BeginExpansion
\begin{shaded}%
%EndExpansion

\begin{proof}
Expand $q_{\pi}^{\tau,\gamma}$ and $q_{\pi}^{\tau+1,\gamma}$,
\begin{align*}
q_{\pi}^{\tau+1,\gamma}-q_{\pi}^{\tau,\gamma} &  =\left(  1-\gamma\right)
\sum_{t=1}^{\tau+1}\gamma^{t-1}q_{\pi}^{t}+\gamma^{\tau+1}q_{\pi}^{\tau+1}\\
&  -\left(  1-\gamma\right)  \sum_{t=1}^{\tau}\gamma^{t-1}q_{\pi}^{t}%
-\gamma^{\tau}q_{\pi}^{\tau}\\
&  =\gamma^{\tau}\left(  q_{\pi}^{\tau+1}-q_{\pi}^{\tau}\right)  \text{.}%
\end{align*}
By definition,%
\begin{align*}
q_{\pi}^{\tau+1}-q_{\pi}^{\tau} &  =\mathbb{E}_{o|ha}\mathbb{E}_{a_{1}%
o_{1}\cdots a_{\tau}o_{\tau}|hao}g\left(  haoa_{1}o_{1}\cdots a_{\tau}o_{\tau
}\Vert h\right)  \\
&  -\mathbb{E}_{o|ha}\mathbb{E}_{a_{1}o_{1}\cdots a_{\tau-1}o_{\tau-1}%
|hao}g\left(  haoa_{1}o_{1}\cdots a_{\tau-1}o_{\tau-1}\Vert h\right)  \\
&  =\mathbb{E}_{o|ha}\mathbb{E}_{a_{1}o_{1}\cdots a_{\tau-1}o_{\tau-1}|hao}\\
&  \left[  \mathbb{E}_{a_{\tau}o_{\tau}|h\cdots o_{\tau-1}}g\left(
haoa_{1}o_{1}\cdots a_{\tau}o_{\tau}\Vert h\right)  -g\left(  haoa_{1}%
o_{1}\cdots a_{\tau-1}o_{\tau-1}\Vert h\right)  \right]  \text{.}%
\end{align*}
Using Eq.\ref{eq-02},
\begin{align*}
&  \mathbb{E}_{a_{\tau}o_{\tau}|h\cdots o_{\tau-1}}g\left(  haoa_{1}%
o_{1}\cdots a_{\tau}o_{\tau}\Vert h\right)  -g\left(  haoa_{1}o_{1}\cdots
a_{\tau-1}o_{\tau-1}\Vert h\right)  \\
&  =\mathbb{E}_{a_{\tau}|h\cdots o_{\tau-1}}g\left(  a_{\tau}\Vert h\cdots
o_{\tau-1}\right)  \text{,}%
\end{align*}
thus%
\begin{align*}
q_{\pi}^{\tau+1,\gamma}-q_{\pi}^{\tau,\gamma} &  =\gamma^{\tau}\mathbb{E}%
_{o|ha}\mathbb{E}_{a_{1}o_{1}\cdots a_{\tau-1}o_{\tau-1}|hao}\mathbb{E}%
_{a_{\tau}|h\cdots o_{\tau-1}}g\left(  a_{\tau}\Vert h\cdots o_{\tau
-1}\right)  \\
&  =\gamma^{\tau}\mathbb{E}_{oa_{1}\cdots o_{\tau-1}a_{\tau}|ha}g\left(
a_{\tau}\Vert h\cdots o_{\tau-1}\right)  \text{.}%
\end{align*}

\end{proof}%

%TCIMACRO{\TeXButton{endshaded}{\end{shaded}}}%
%BeginExpansion
\end{shaded}%
%EndExpansion

\begin{lemma}
\label{lem-02}$q^{\tau+1,\gamma}\left(  h,a\right)  -q^{\tau,\gamma}\left(
h,a\right)  \leq\gamma^{\tau}\mathbb{E}_{o|ha}\max_{a_{1}}\mathbb{E}%
_{o_{1}|haoa_{1}}\cdots\max_{a_{\tau}}g\left(  a_{\tau}\Vert h\cdots
o_{\tau-1}\right)  $.
\end{lemma}

%

%TCIMACRO{\TeXButton{beginshaded}{\begin{shaded}}}%
%BeginExpansion
\begin{shaded}%
%EndExpansion

\begin{proof}
Expand $q^{\tau,\gamma}$ and $q^{\tau+1,\gamma}$, and note that $\max X-\max
Y\leq\max\left\vert X-Y\right\vert $, then%
\begin{align*}
&  q^{\tau+1,\gamma}\left(  h,a\right)  -q^{\tau,\gamma}\left(  h,a\right)  \\
&  =\mathbb{E}_{o|ha}\max_{a_{1}}\mathbb{E}_{o_{1}|haoa_{1}}\cdots
\max_{a_{\tau}}\left[  g\left(  a\Vert h\right)  +\gamma g\left(  a_{1}\Vert
hao\right)  +\cdots+\gamma^{\tau}g\left(  a_{\tau}\Vert h\cdots o_{\tau
-1}\right)  \right]  \\
&  -\mathbb{E}_{o|ha}\max_{a_{1}}\mathbb{E}_{o_{1}|haoa_{1}}\cdots
\max_{a_{\tau-1}}\left[  g\left(  a\Vert h\right)  +\gamma g\left(  a_{1}\Vert
hao\right)  +\cdots+\gamma^{\tau-1}g\left(  a_{\tau-1}\Vert h\cdots o_{\tau
-2}\right)  \right]  \\
&  \leq\mathbb{E}_{o|ha}\max_{a_{1}}\{\mathbb{E}_{o_{1}|haoa_{1}}\cdots
\max_{a_{\tau}}\left[  g\left(  a\Vert h\right)  +\gamma g\left(  a_{1}\Vert
hao\right)  +\cdots+\gamma^{\tau}g\left(  a_{\tau}\Vert h\cdots o_{\tau
-1}\right)  \right]  \\
&  -\mathbb{E}_{o_{1}|haoa_{1}}\cdots\max_{a_{\tau-1}}\left[  g\left(  a\Vert
h\right)  +\gamma g\left(  a_{1}\Vert hao\right)  +\cdots+\gamma^{\tau
-1}g\left(  a_{\tau-1}\Vert h\cdots o_{\tau-2}\right)  \right]  \}\\
&  \leq\cdots\\
&  \leq\gamma^{\tau}\mathbb{E}_{o|ha}\max_{a_{1}}\mathbb{E}_{o_{1}|haoa_{1}%
}\cdots\max_{a_{\tau}}g\left(  a_{\tau}\Vert h\cdots o_{\tau-1}\right)
\text{.}%
\end{align*}

\end{proof}%

%TCIMACRO{\TeXButton{endshaded}{\end{shaded}}}%
%BeginExpansion
\end{shaded}%
%EndExpansion

It can be seen that if $\mathbb{E}_{oa_{1}\cdots o_{\tau-1}a_{\tau}%
|ha}g\left(  a_{\tau}\Vert h\cdots o_{\tau-1}\right)  $ is bounded, then both
$q_{\pi}^{\gamma,\tau}$ and $q^{\gamma,\tau}$ are a Cauchy sequences with
respect to $\tau$.

\section{Exploration in Finite Markovian Environment with Dirichlet Priors}

In this section we restrict the discussion to a simple case, where the
possible actions and sensory inputs are finite, and the agent assumes that the
environment is Markovian, namely, the current sensory input and action is
sufficient for determining the (probabilities of the) next sensory inputs.
Formally, let $\mathcal{S}=\left\{  1,\cdots,S\right\}  $ and $\mathcal{A}%
=\left\{  1,\cdots,A\right\}  $ be the space of possible sensory inputs, to
which we referred as `states', and actions. The dynamics of the environment is
fully determined by the transition probability $p\left(  s^{\prime
}|s,a\right)  $.

The agent tries to learn the transition probabilities. Initially, it assumes
for each $\left\langle s,a\right\rangle $ a Dirichlet prior over the random
variable $\Theta_{s,a}$ corresponding to the distribution $p\left(
\cdot|s,a\right)  $. Through time, the agent observes the transitions when
performing $a$ at $s$, and updates its estimate of $\Theta_{s,a}$ through
probabilistic inference. Since the Dirichlet distribution is conjugate with
multinomial distribution, the posterior is still a Dirichlet distribution over
$\Theta_{s,a}$. Therefore, at any time, the agent's knowledge about the
environment can be fully summarized by a three dimensional array
$\alpha\left(  s,a,s^{\prime}\right)  $, such that $Dir\left(  \alpha
_{s,a,1},\cdots,\alpha_{s,a,S}\right)  $ is the current (prior or posterior)
density of $\Theta_{s,a}$, and the definition of the optimal curiosity Q-value
can be written as\footnote{We use $\alpha_{s,a}$ both for the vector $\left[
\alpha_{s,a,1},\cdots,\alpha_{s,a,S}\right]  $ and the number $\sum
_{s^{\prime}}\alpha_{s,a,s^{\prime}}$. The meaning should be clear from the
context.}%
\begin{align*}
q_{\alpha}^{\gamma,\tau}\left(  s,a\right)   &  =g\left(  \alpha_{s,a}\right)
+\gamma\sum_{s^{\prime}}p_{\alpha}\left(  s^{\prime}|s,a\right)
\max_{a^{\prime}}q_{\alpha\triangleleft\left\langle s,a,s^{\prime
}\right\rangle }^{\gamma,\tau-1}\left(  s^{\prime},a^{\prime}\right) \\
&  =g\left(  \alpha_{s,a}\right)  +\gamma\sum_{s^{\prime}}\frac{\alpha
_{s,a,s^{\prime}}}{\alpha_{s,a}}\max_{a^{\prime}}q_{\alpha\triangleleft
\left\langle s,a,s^{\prime}\right\rangle }^{\gamma,\tau-1}\left(  s^{\prime
},a^{\prime}\right)  \text{.}%
\end{align*}
Here $g\left(  \alpha_{s,a}\right)  $ is the expected immediate information
gain for $\Theta_{s,a}$ given the current parameterization of the Dirichlet
distribution. By definition, $g\left(  \alpha_{s,a}\right)  $ is also the
mutual information between $\Theta$ and an additional observation. According
to \cite{01-KLDir}, the precise form of $g$ is given by%
\[
g\left(  \left[  n_{1},\cdots,n_{S}\right]  \right)  =\log n^{\ast}%
-\psi\left(  n^{\ast}+1\right)  -\sum_{s=1}^{S}\frac{n_{s}}{n^{\ast}}\left[
\log n_{s}-\psi\left(  n_{s}+1\right)  \right]  \text{,}%
\]
where $n^{\ast}=\sum_{s}n_{s}$, and $\psi\left(  \cdot\right)  $ is the
standard digamma function. By marginalizing out $\Theta_{s,a}$, the predictive
probability is given by $p_{\alpha}\left(  s^{\prime}|s,a\right)
=\frac{\alpha_{s,a,s^{\prime}}}{\alpha_{s,a}}$, and $\triangleleft$ is the
operator\footnote{Assume that the operator $\triangleleft$ is right
associative, so one can write $\alpha\triangleleft\left\langle s_{1}%
,a_{2},s_{2}\right\rangle \triangleleft\left\langle s_{2},a_{3},s_{3}%
\right\rangle \cdots$, or simply $\alpha\triangleleft\left\langle s_{1}%
a_{2}s_{2}a_{3}s_{3}\cdots\right\rangle $.} such that $\alpha\triangleleft
\left\langle s,a,s^{\prime}\right\rangle $ is the same as $\alpha$, except
that the entry indexed by $\left\langle s,a,s^{\prime}\right\rangle $ is
increased by $1$. Similarly, for a given policy $\pi$, the curiosity Q-value
can be written as%
\[
q_{\pi,\alpha}^{\gamma,\tau}\left(  s,a\right)  =g\left(  \alpha_{s,a}\right)
+\gamma\sum_{s^{\prime}}\frac{\alpha_{s,a,s^{\prime}}}{\alpha_{s,a}}%
\sum_{a^{\prime}}\pi_{\alpha}\left(  a^{\prime}|s^{\prime}\right)
q_{\pi,\alpha\triangleleft\left\langle s,a,s^{\prime}\right\rangle }%
^{\gamma,\tau-1}\left(  s^{\prime},a^{\prime}\right)  \text{.}%
\]

\subsection{Curiosity Q-value in Infinite Time Horizon}

In this subsection we extend the definition of curiosity Q-value to infinite
horizon. We show that a) the limit $\lim_{\tau\rightarrow\infty}q_{\pi,\alpha
}^{\gamma,\tau}$ exists, b) the limit $\lim_{\tau\rightarrow\infty}q_{\alpha
}^{\gamma,\tau}$ exists, and c) the limit is the solution of the infinite recursion.

\begin{proposition}
\label{prop-01}$q_{\pi,\alpha}^{\gamma}\left(  s,a\right)  =\lim
_{\tau\rightarrow\infty}q_{\pi,\alpha}^{\gamma,\tau}\left(  s,a\right)  $
exists for any $\pi$, $\alpha$, $s$, $a$, and $\gamma\in\left[  0,1\right)  $.
Moreover, the convergence is uniform with respect to $\left\langle
s,a\right\rangle $ in the sense that%
\[
0\leq q_{\pi,\alpha}^{\gamma}\left(  s,a\right)  -q_{\pi,\alpha}^{\gamma,\tau
}\left(  s,a\right)  \leq\frac{g_{\alpha}}{1-\gamma}\gamma^{\tau}\text{,
}\forall s,a
\]
where $g_{\alpha}=\max_{s}\max_{a}g\left(  \alpha_{s,a}\right)  $.
\end{proposition}

%

%TCIMACRO{\TeXButton{beginshaded}{\begin{shaded}}}%
%BeginExpansion
\begin{shaded}%
%EndExpansion

\begin{proof}
Rewrite the result in Lemma.\ref{lem-01} in this context:%
\begin{align*}
q_{\pi,\alpha}^{\gamma,\tau+1}\left(  s,a\right)  -q_{\pi,\alpha}^{\gamma
,\tau}\left(  s,a\right)   &  =\gamma^{\tau}\mathbb{E}_{oa_{1}\cdots
o_{\tau-1}a_{\tau}|ha}g\left(  a_{\tau}\Vert h\cdots o_{\tau-1}\right)  \\
&  =\gamma^{\tau}\mathbb{E}_{s_{1}a_{2}s_{2}\cdots s_{\tau}a_{\tau+1}%
|ha}g\left(  \alpha_{s_{\tau}a_{\tau+1}}^{\prime}\right)  \text{,}%
\end{align*}
where%
\[
\alpha^{\prime}=\alpha\triangleleft\left\langle s,a,s_{1}\right\rangle
\triangleleft\left\langle s_{1},a_{2},s_{2}\right\rangle \triangleleft
\cdots\triangleleft\left\langle s_{\tau-1},a_{\tau},s_{\tau}\right\rangle
\text{.}%
\]
Because $g\left(  \alpha_{s,a}^{\prime}\right)  $ depends only on the
transitions when performing $a$ at $s$, and all such transitions are
exchangeable since they are assumed to be i.i.d. when $\Theta_{s,a}$ is given,
one can rewrite the expectation in the following form:%
\[
\mathbb{E}_{s_{1}a_{2}s_{2}\cdots s_{\tau}a_{\tau+1}|ha}g\left(
\alpha_{s_{\tau}a_{\tau+1}}^{\prime}\right)  =\mathbb{E}_{s}\mathbb{E}%
_{a}\mathbb{E}_{n}\mathbb{E}_{x_{1}}\cdots\mathbb{E}_{x_{n}}g\left(
\alpha_{s,a}^{\prime}\right)  \text{.}%
\]
The first and second expectations are taken over the possible final
state-action pairs $\left\langle s_{\tau},a_{\tau+1}\right\rangle $, from
which $g\left(  \alpha_{s_{\tau}a_{\tau+1}}^{\prime}\right)  $ is computed.
Once $\left\langle s,a\right\rangle $ is fixed, the third expectation is taken
over the time $n$ that $\left\langle s,a\right\rangle $-pair appears in the
trajectory $sas_{1}a_{2}\cdots s_{\tau}$, i.e., the time that transitions
starting from $s$ with action $a$ occurs. The rest of the expectations are
over the $n$ destinations of the transitions, denoted as $x_{1},\cdots,x_{n}$.
By definition, $Dir\left(  \alpha_{s,a}^{\prime}\right)  $ is the posterior
distribution after seeing $x_{1},\cdots,x_{n}$, and $g\left(  \alpha
_{s,a}^{\prime}\right)  $ is the expected information gain of seeing the
outcome of the $\left(  n+1\right)  $-th transition, which we denote $x_{n+1}%
$, thus%
\[
g\left(  \alpha_{s,a}^{\prime}\right)  =I\left(  \Theta_{s,a};X_{n+1}%
|x_{1},\cdots,x_{n}\right)  \text{,}%
\]
and%
\[
\mathbb{E}_{x_{1}}\cdots\mathbb{E}_{x_{n}}g\left(  \alpha_{s,a}^{\prime
}\right)  =I\left(  \Theta_{s,a};X_{n+1}|X_{1},\cdots,X_{n}\right)  \text{.}%
\]
Note that $X_{1},\cdots,X_{n+1}$ are i.i.d. given $\Theta_{s,a}$, therefore%
\begin{align*}
&  I\left(  \Theta_{s,a};X_{n+1}|X_{1},\cdots,X_{n}\right)  \\
&  =I\left(  \Theta_{s,a};X_{1},\cdots,X_{n+1}\right)  -I\left(  \Theta
_{s,a};X_{1},\cdots,X_{n}\right)  \\
&  =H\left(  X_{1},\cdots,X_{n+1}\right)  -\sum_{i=1}^{n+1}H\left(
X_{i}|\Theta\right)  -H\left(  X_{1},\cdots,X_{n}\right)  +\sum_{i=1}%
^{n}H\left(  X_{i}|\Theta\right)  \\
&  =H\left(  X_{n+1}|X_{1},\cdots,X_{n}\right)  -H\left(  X_{n+1}%
|\Theta\right)  \\
&  \leq H\left(  X_{n+1}\right)  -H\left(  X_{n+1}|\Theta\right)  \\
&  =I\left(  \Theta;X_{n+1}\right)  \\
&  =I\left(  \Theta;X_{1}\right)  \text{.}%
\end{align*}
This means that $I\left(  \Theta_{s,a};X_{n+1}|X_{1},\cdots,X_{n}\right)  $ is
upper bounded by $I\left(  \Theta;X_{1}\right)  $, which is the expected
information gain of seeing the outcome of the transition for the first time.
By definition $I\left(  \Theta;X_{1}\right)  =g\left(  \alpha_{s,a}\right)  $,
and it follows that%
\[
\mathbb{E}_{n}\mathbb{E}_{x_{1}}\cdots\mathbb{E}_{x_{n}}g\left(  \alpha
_{s,a}^{\prime}\right)  \leq g\left(  \alpha_{s,a}\right)  \text{.}%
\]
Therefore,%
\begin{align*}
q_{\pi,\alpha}^{\gamma,\tau+1}\left(  s,a\right)  -q_{\pi,\alpha}^{\gamma
,\tau}\left(  s,a\right)   &  =\gamma^{\tau}\mathbb{E}_{s_{1}a_{2}s_{2}\cdots
s_{\tau}a_{\tau+1}|ha}g\left(  \alpha_{s_{\tau}a_{\tau+1}}^{\prime}\right)  \\
&  =\gamma^{\tau}\mathbb{E}_{s}\mathbb{E}_{a}\mathbb{E}_{n}\mathbb{E}_{x_{1}%
}\cdots\mathbb{E}_{x_{n}}g\left(  \alpha_{s,a}^{\prime}\right)  \\
&  \leq\gamma^{\tau}\mathbb{E}_{s}\mathbb{E}_{a}g\left(  \alpha_{s,a}\right)
\\
&  \leq\gamma^{\tau}\max_{s}\max_{a}g\left(  \alpha_{s,a}\right)  \\
&  \leq\gamma^{\tau}g_{\alpha}\text{.}%
\end{align*}
Since $g_{\alpha}$ depends on $\alpha$ only, for any $T$
\[
q_{\pi,\alpha}^{\gamma,\tau+T}\left(  s,a\right)  -q_{\pi,\alpha}^{\gamma
,\tau}\left(  s,a\right)  \leq\frac{g_{\alpha}}{1-\gamma}\gamma^{\tau}\text{.}%
\]
This means that $q_{\pi,\alpha}^{\gamma,\tau}\left(  s,a\right)  $ is a Cauchy
sequence with respect to $\tau$, thus $\lim_{\tau\rightarrow\infty}%
q_{\pi,\alpha}^{\gamma,\tau}\left(  s,a\right)  $ exists. Also note that the
convergence is uniform since $g_{\alpha}$ does not depend on $\left\langle
s,a\right\rangle $.
\end{proof}

%

%TCIMACRO{\TeXButton{endshaded}{\end{shaded}}}%
%BeginExpansion
\end{shaded}%
%EndExpansion

\begin{proposition}
\label{prop-02}$q_{\alpha}^{\gamma}\left(  s,a\right)  =\lim_{\tau
\rightarrow\infty}q_{\alpha}^{\gamma,\tau}\left(  s,a\right)  $ exists for any
$\alpha$, $s$, $a$, and $\gamma\in\left[  0,1\right)  $. Also the convergence
is uniform in the sense that%
\[
0\leq q^{\gamma}\left(  s,a\right)  -q_{\alpha}^{\gamma,\tau}\left(
s,a\right)  \leq\frac{g_{\alpha}}{1-\gamma}\gamma^{\tau}\text{.}%
\]
\end{proposition}%

%TCIMACRO{\TeXButton{beginshaded}{\begin{shaded}}}%
%BeginExpansion
\begin{shaded}%
%EndExpansion

\begin{proof}
Rewrite the result in Lemma.\ref{lem-02},%
\begin{align*}
q_{\alpha}^{\gamma,\tau+1}\left(  s,a\right)  -q_{\alpha}^{\gamma,\tau}\left(
s,a\right)   &  \leq\gamma^{\tau}\mathbb{E}_{s_{1}|sa}\max_{a_{2}}%
\mathbb{E}_{s_{2}|s_{1}a_{2},\alpha\triangleleft\left\langle s,a,s_{1}%
\right\rangle }\\
&  \cdots\mathbb{E}_{s_{\tau}|s_{\tau-1}a_{\tau},\alpha\triangleleft
\left\langle sas_{1}\cdots s_{\tau-1}\right\rangle }\max_{a_{\tau+1}}g\left(
\alpha_{s_{\tau},\alpha_{\tau+1}}^{\prime}\right)  \text{.}%
\end{align*}
Since the $\max$ operator is only over actions, the proof in the previous
proposition still holds, so%
\[
q_{\alpha}^{\gamma,\tau+1}\left(  s,a\right)  -q_{\alpha}^{\gamma,\tau}\left(
s,a\right)  \leq\gamma^{\tau}g_{\alpha}\text{,}%
\]
and the result follows.
\end{proof}

%

%TCIMACRO{\TeXButton{endshaded}{\end{shaded}}}%
%BeginExpansion
\end{shaded}%
%EndExpansion

The next proposition shows that $q_{\alpha}^{\gamma}$ is the solution to the
infinite recursion.

\begin{proposition}
\label{prop-03}$q_{\alpha}^{\gamma}$ is the solution to the recursion%
\[
q_{\alpha}^{\gamma}\left(  s,a\right)  =g\left(  \alpha_{s,a}\right)
+\gamma\sum_{s^{\prime}}\frac{\alpha_{s,a,s^{\prime}}}{\alpha_{s,a}}%
\max_{a^{\prime}}q_{\alpha\triangleleft\left\langle s,a,s^{\prime
}\right\rangle }^{\gamma}\left(  s^{\prime},a^{\prime}\right)  \text{,}%
\]
and for any other policy $\pi$, $q_{\alpha}^{\gamma}\left(  s,a\right)  \geq
q_{\pi,\alpha}^{\gamma}\left(  s,a\right)  $.
\end{proposition}

%

%TCIMACRO{\TeXButton{beginshaded}{\begin{shaded}}}%
%BeginExpansion
\begin{shaded}%
%EndExpansion

\begin{proof}
To see that $q_{\alpha}^{\gamma}$ is the solution, taking any $\varepsilon>0$,
one can find a $\tau$ such that $\left\vert q_{\alpha}^{\gamma,\tau
+1}-q_{\alpha}^{\gamma}\right\vert <\frac{\varepsilon}{2}$, and $\left\vert
q_{\alpha\triangleleft\left\langle s,a,s^{\prime}\right\rangle }^{\gamma,\tau
}-q_{\alpha\triangleleft\left\langle s,a,s^{\prime}\right\rangle }^{\gamma
}\right\vert <\frac{\varepsilon}{2}$ for any $\left\langle s,a,s^{\prime
}\right\rangle $, thanks to the fact that there are only finite number of
$\left\langle s,a,s^{\prime}\right\rangle $ triples, and the convergence from
$q_{\alpha}^{\gamma,\tau+1}\left(  s,a\right)  $ to $q_{\alpha}^{\gamma
}\left(  s,a\right)  $ is uniform. It follows that%
\begin{align*}
&  \left\vert g\left(  \alpha_{s,a}\right)  +\gamma\sum_{s^{\prime}}%
\frac{\alpha_{s,a,s^{\prime}}}{\alpha_{s,a}}\max_{a^{\prime}}q_{\alpha
\triangleleft\left\langle s,a,s^{\prime}\right\rangle }^{\gamma}\left(
s^{\prime},a^{\prime}\right)  -q_{\alpha}^{\gamma}\left(  s,a\right)
\right\vert \\
&  \leq\left\vert q_{\alpha}^{\gamma,\tau+1}-q_{\alpha}^{\gamma}\right\vert
+\gamma\left\vert q_{\alpha\triangleleft\left\langle s,a,s^{\prime
}\right\rangle }^{\gamma,\tau}-q_{\alpha\triangleleft\left\langle
s,a,s^{\prime}\right\rangle }^{\gamma}\right\vert \\
&  <\varepsilon\text{.}%
\end{align*}
Since $\varepsilon$ is chosen arbitrary, $q_{\alpha}^{\gamma}\left(
s,a\right)  $ must be the solution of the infinite recursion.

The fact that $q_{\alpha}^{\gamma}\left(  s,a\right)  \geq q_{\pi,\alpha
}^{\gamma}\left(  s,a\right)  $ follows from the fact that $q_{\alpha}%
^{\gamma,\tau}$ and $q_{\pi,\alpha}^{\gamma,\tau}$ are monotonically
increasing on $\tau$ (by Lemma.\ref{lem-01}), and $q_{\alpha}^{\gamma,\tau
}\geq q_{\pi,\alpha}^{\gamma,\tau}$ for any given $\tau$ and $\pi$.
\end{proof}

%

%TCIMACRO{\TeXButton{endshaded}{\end{shaded}}}%
%BeginExpansion
\end{shaded}%
%EndExpansion

The propositions above guarantees the existence and optimality of $q_{\alpha
}^{\gamma}$, and the following discussions would focus on $q_{\alpha}^{\gamma
}$. We drop the super-script $\gamma$ in the rest of this section.

\subsection{Approximation through Dynamic Programming}

The optimal curiosity Q-value is given by the infinite recursion%
\begin{equation}
q_{\alpha}\left(  s,a\right)  =g\left(  \alpha_{s,a}\right)  +\gamma
\sum_{s^{\prime}}\frac{\alpha_{s,a,s^{\prime}}}{\alpha_{s,a}}\max_{a^{\prime}%
}q_{\alpha\triangleleft\left\langle s,a,s^{\prime}\right\rangle }\left(
s^{\prime},a^{\prime}\right)  \text{.} \label{eq-05}%
\end{equation}
It is impossible to solve this equation directly. A natural idea is to
approximate this infinite recursion by solving at each time step the following
Bellman equation,%
\begin{equation}
\tilde{q}_{\alpha}\left(  s,a\right)  =g\left(  \alpha_{s,a}\right)
+\gamma\sum_{s^{\prime}}\frac{\alpha_{s,a,s^{\prime}}}{\alpha_{s,a}}%
\max_{a^{\prime}}\tilde{q}_{\alpha}\left(  s^{\prime},a^{\prime}\right)
\text{.} \label{eq-06}%
\end{equation}
The Bellman equation can be solved by dynamic programming in time polynomial
on $S$ and $A$. The algorithm is given below.

\bigskip

\begin{center}%
\begin{tabular}
[c]{rl}\hline
\multicolumn{2}{l}{\texttt{CuriosityExploreDP}$\left(  \alpha,\gamma
,s,T\right)  $}\\
\multicolumn{2}{l}{Input: prior $\alpha$, discount factor $\gamma$, initial
state $s$, number of steps $T$.}\\
\multicolumn{2}{l}{Output: posterior stored in $\alpha$}\\\hline\hline
1 & $G\leftarrow\left[  g\left(  \alpha_{s,a}\right)  \right]  _{\left\langle
s,a\right\rangle \in\mathcal{S}\times\mathcal{A}}$\\
2 & $P\leftarrow\left[  \frac{\alpha_{s,a,s^{\prime}}}{\alpha_{s,a}}\right]
_{\left\langle s,a,s^{\prime}\right\rangle \in\mathcal{S}\times\mathcal{A}%
\times\mathcal{S}}$\\
3 & \textbf{for} $t$ \textbf{in} $1$ \textbf{to} $T$\\
4 & \ $\pi\leftarrow$\texttt{PolicyIteration}$\left(  G,P,\gamma\right)  $\\
5 & \ $a\leftarrow\pi_{s}$\\
6 & \ $s^{\prime}\leftarrow$\texttt{NextState}$\left(  a\right)  $\\
7 & \ $\alpha_{s,a,s^{\prime}}\leftarrow\alpha_{s,a,s^{\prime}}+1$\\
8 & \ $G_{s,a}\leftarrow g\left(  \alpha_{s,a}\right)  $\\
9 & \ \textbf{for} $s^{\prime}$ \textbf{in} $1$ \textbf{to} $S$\\
10 & \ \ \ $P_{s,a,s^{\prime}}\leftarrow\frac{\alpha_{s,a,s^{\prime}}}%
{\alpha_{s,a}}$\\
11 & \ \textbf{end} \textbf{for}\\
12 & \ $s\leftarrow s^{\prime}$\\
13 & \textbf{end} \textbf{for}\\\hline
\end{tabular}

\bigskip
\end{center}

A surprising fact is that when $\alpha$ is large, $\tilde{q}_{\alpha}$ is in
fact a very good approximation of $q_{\alpha}$, which is the central result in
this section. We start by investigating the properties of the gain $g\left(
\alpha_{s,a}^{t}\right)  $.

\subsection{Properties of Expected Information Gain in Dirichlet Case}

The expected information gain of a Dirichlet distribution $Dir\left(
n_{1},\cdots,n_{S}\right)  $ is given by%
\[
g\left(  n\right)  =\log\left(  n\right)  -\psi\left(  n+1\right)  -\sum
_{s}\frac{n_{s}}{n}\left[  \log\left(  n_{s}\right)  -\psi\left(
n_{s}+1\right)  \right]  \text{.}%
\]
Define%
\begin{align*}
f\left(  x\right)   &  =x\left[  \psi\left(  x+1\right)  -\log x\right] \\
&  =1-x\left[  \log x-\psi\left(  x\right)  \right]  \text{.}%
\end{align*}
then%
\[
g\left(  n\right)  =\frac{1}{n}\left[  \sum_{s}f\left(  n_{s}\right)
-f\left(  n\right)  \right]  \text{.}%
\]

The following important properties has been proved by Alzer in
\cite{97-IneqPsi}.

\begin{theorem}
\label{theo-01}$f$ has the following properties\footnote{Alzer's original
paper considers the function $x\left(  \log x-\psi\left(  x\right)  \right)
=1-f\left(  x\right)  $. Here the statements are modified accordingly.}:

\begin{enumerate}
\item[a)] $\lim_{x\rightarrow0}f\left(  x\right)  =0$, $\lim_{x\rightarrow
\infty}f\left(  x\right)  =\frac{1}{2}$

\item[b)] $f$ is strictly completely monotonic, in the sense that%
\[
\left(  -1\right)  ^{n+1}\frac{d^{n}f\left(  x\right)  }{dx^{n}}>0\text{.}%
\]

\end{enumerate}
\end{theorem}

In particular, Theorem.\ref{theo-01} shows that $f$ is strictly monotonically
increasing, and also strictly concave. The following Lemma summarizes the
properties about $f$ used in this paper.

\begin{lemma}
\label{lem-g1}Define $\delta_{m}\left(  x\right)  =f\left(  x+m\right)
-f\left(  x\right)  $ for $m>0$. Then

\begin{enumerate}
\item[a)] $f$ is sub-additive, i.e., $f\left(  x\right)  +f\left(  y\right)
>f\left(  x+y\right)  $ for $x,y>0$

\item[b)] $\delta_{m}\left(  x\right)  $ is monotonically decreasing on
$\left(  0,\infty\right)  $.

\item[c)] $0<x\delta_{m}\left(  x\right)  <\frac{1-e^{-1}}{2}m$ for
$x\in\left(  0,\infty\right)  $.
\end{enumerate}
\end{lemma}

%

%TCIMACRO{\TeXButton{beginshaded}{\begin{shaded}}}%
%BeginExpansion
\begin{shaded}%
%EndExpansion

\begin{proof}
a) Note that $g\left(  n\right)  $ is mutual information, and the unknown
observation depends on the parameters of the distribution, therefore $g\left(
n\right)  >0$, and%
\[
0<g\left(  \left[  x,y\right]  \right)  =\frac{1}{x+y}\left[  f\left(
x\right)  +f\left(  y\right)  -f\left(  x+y\right)  \right]  \text{.}%
\]

b) Note that%
\[
\delta_{m}\left(  x\right)  =\int_{0}^{m}f^{\prime}\left(  x+s\right)
ds\text{,}%
\]
and the result follows from $f^{\prime\prime}\left(  x\right)  <0$.

c) Clearly, $x\delta_{m}\left(  x\right)  >0$ because $f$ is strictly
increasing. From Intermediate Value Theorem, there some $\delta\in\left(
0,m\right)  $, such that%
\begin{align*}
x\delta_{m}\left(  x\right)   &  =x\left[  f\left(  x+m\right)  -f\left(
x\right)  \right] \\
&  =mxf^{\prime}\left(  x+\delta\right) \\
&  =mxf^{\prime}\left(  x\right)  +mx\left[  f^{\prime}\left(  x+\delta
\right)  -f^{\prime}\left(  x\right)  \right] \\
&  <mxf^{\prime}\left(  x\right)  \text{.}%
\end{align*}
The inequality is because $f$ is strictly concave.

From \cite{97-IneqPsi},
\[
f\left(  x\right)  =1-x\int_{0}^{\infty}\phi\left(  t\right)  e^{-tx}%
dt\text{,}%
\]
where%
\[
\phi\left(  t\right)  =\frac{1}{1-e^{-t}}-\frac{1}{t}%
\]
is strictly increasing, with $\lim_{t\rightarrow0}\phi\left(  t\right)
=\frac{1}{2}$ and $\lim_{t\rightarrow\infty}\phi\left(  t\right)  =1$.
Therefore,%
\begin{align*}
xf^{\prime}\left(  x\right)   &  =x\int_{0}^{\infty}\phi\left(  t\right)
e^{-tx}\left(  xt-1\right)  dt\\
&  =x^{2}\int_{0}^{\frac{1}{x}}\phi\left(  t\right)  e^{-tx}\left(  t-\frac
{1}{x}\right)  dt+x^{2}\int_{\frac{1}{x}}^{\infty}\phi\left(  t\right)
e^{-tx}\left(  t-\frac{1}{x}\right)  dt\\
&  <x^{2}\phi\left(  0\right)  \int_{0}^{\frac{1}{x}}e^{-tx}\left(  t-\frac
{1}{x}\right)  dt+x^{2}\phi\left(  \infty\right)  \int_{\frac{1}{x}}^{\infty
}e^{-tx}\left(  t-\frac{1}{x}\right)  dt\\
&  =\frac{x^{2}}{2}\int_{0}^{\infty}e^{-tx}\left(  t-\frac{1}{x}\right)
dt+\frac{x^{2}}{2}\int_{\frac{1}{x}}^{\infty}e^{-tx}tdt-\frac{x}{2}\int
_{\frac{1}{x}}^{\infty}e^{-tx}dt\\
&  <\frac{x^{2}}{2}\int_{0}^{\infty}e^{-tx}\left(  t-\frac{1}{x}\right)
dt+\frac{x^{2}}{2}\int_{0}^{\infty}e^{-tx}tdt-\frac{x}{2}\int_{\frac{1}{x}%
}^{\infty}e^{-tx}dt\text{.}%
\end{align*}
Note that%
\[
\int_{0}^{\infty}e^{-tx}tdt=\frac{1}{x^{2}}\text{, }\int_{0}^{\infty}%
e^{-tx}dt=\frac{1}{x}\text{,}%
\]
and%
\[
x\int_{\frac{1}{x}}^{\infty}e^{-tx}dt=e^{-1}\text{,}%
\]
it follows that%
\[
xf^{\prime}\left(  x\right)  <\frac{1-e^{-1}}{2}\text{.}%
\]

\end{proof}%

%TCIMACRO{\TeXButton{endshaded}{\end{shaded}}}%
%BeginExpansion
\end{shaded}%
%EndExpansion

The properties of $f$ guarantee that $g\left(  n\right)  $ decreases at the
rate of $\frac{1}{n}$. The result is formulated in the following Lemma.

\begin{lemma}
\label{lem-g2}Let $Dir\left(  n_{1}^{0},\cdots,n_{S}^{0}\right)  $ and
$Dir\left(  n_{1}^{t},\cdots,n_{S}^{t}\right)  $ be the prior and the
posterior distribution, such that $n^{t}=n^{0}+t$. Let $s^{\ast}=\arg\max
_{s}n_{s}^{0}$. Then%
\[
\frac{\sum_{s\neq s^{\ast}}f\left(  n_{s}^{0}\right)  -f\left(  \sum_{s\neq
s^{\ast}}n_{s}^{0}\right)  }{2n}<g\left(  n^{t}\right)  <\frac{S-1}%
{2n}\text{.}%
\]

\end{lemma}%

%TCIMACRO{\TeXButton{beginshaded}{\begin{shaded}}}%
%BeginExpansion
\begin{shaded}%
%EndExpansion

\begin{proof}
The upper bound is because $0<f\left(  x\right)  <\frac{1}{2}$ and $f$ is
increasing, thus%
\begin{align*}
\sum_{s}f\left(  n_{s}^{t}\right)  -f\left(  n^{t}\right)   &  =f\left(
n_{1}^{t}\right)  -f\left(  n^{t}\right)  +\sum_{s\neq1}f\left(  n_{s}%
^{t}\right) \\
&  <\frac{S-1}{2}\text{.}%
\end{align*}
The lower bound follows from the fact that $f\left(  x+m\right)  -f\left(
x\right)  $ is decreasing. We show that the trajectory minimizing $g\left(
n^{t}\right)  $ is the one such that all $t$ observations equal to $s^{\ast}$.
To see this, let $m_{s}$ be the number of times observing $s\neq s^{\ast}$,
then%
\begin{align*}
f\left(  n_{s}^{0}+m_{s}\right)  +f\left(  n_{s^{\ast}}^{0}+m_{s^{\ast}%
}\right)   &  =f\left(  n_{s}^{0}\right)  +f\left(  n_{s^{\ast}}%
^{0}+m_{s^{\ast}}+m_{s}\right) \\
&  +f\left(  n_{s}^{0}+m_{s}\right)  -f\left(  n_{s}^{0}\right) \\
&  -\left(  f\left(  n_{s^{\ast}}^{0}+m_{s^{\ast}}+m_{s}\right)  -f\left(
n_{s^{\ast}}^{0}+m_{s^{\ast}}\right)  \right)  \text{.}%
\end{align*}
Note that $n_{s^{\ast}}^{0}+m_{s^{\ast}}\geq n_{s}^{0}$, so%
\[
f\left(  n_{s}^{0}+m_{s}\right)  +f\left(  n_{s^{\ast}}^{0}+m_{s^{\ast}%
}\right)  \geq f\left(  n_{s}^{0}\right)  +f\left(  n_{s^{\ast}}%
^{0}+m_{s^{\ast}}+m_{s}\right)  \text{.}%
\]

Now assume the observations are all $s^{\ast}$, from sub-additivity,%
\begin{align*}
&  \sum_{s}f\left(  n_{s}^{t}\right)  -f\left(  n^{t}\right) \\
&  =\sum_{s\neq s^{\ast}}f\left(  n_{s}^{0}\right)  +f\left(  n_{s^{\ast}}%
^{0}+t\right)  -f\left(  n^{0}+t\right) \\
&  =\sum_{s\neq s^{\ast}}f\left(  n_{s}^{0}\right)  -f\left(  \sum_{s\neq
s^{\ast}}n_{s}^{0}\right)  +\left[  f\left(  \sum_{s\neq s^{\ast}}n_{s}%
^{0}\right)  +f\left(  n_{s^{\ast}}^{0}+t\right)  -f\left(  n^{0}+t\right)
\right] \\
&  >\sum_{s\neq s^{\ast}}f\left(  n_{s}^{0}\right)  -f\left(  \sum_{s\neq
s^{\ast}}n_{s}^{0}\right)  \text{.}%
\end{align*}

\end{proof}%

%TCIMACRO{\TeXButton{endshaded}{\end{shaded}}}%
%BeginExpansion
\end{shaded}%
%EndExpansion

A little remark: The bounds hold irrespective of the data generating process,
namely, it holds for \emph{any} sequences of observations, \emph{including
sequences with zero probabilities}.

The following Lemma bound the variation of the expected information gain, when
one single observation is added.

\begin{lemma}
\label{lem-g3}Let $n=\left[  n_{1},\cdots,n_{S}\right]  $ and $n^{\prime
}=\left[  n_{1},\cdots,n_{s-1},n_{s}+1,n_{s+1},\cdots,n_{S}\right]  $, then%
\[
\frac{n_{s}}{n}\left\vert g\left(  n^{\prime}\right)  -g\left(  n\right)
\right\vert \leq\frac{S}{2n^{2}}\text{, }\forall n_{s}>0\text{.}%
\]

\end{lemma}%

%TCIMACRO{\TeXButton{beginshaded}{\begin{shaded}}}%
%BeginExpansion
\begin{shaded}%
%EndExpansion

\begin{proof}
Without loss of generality let $s=1$. Note that%
\begin{align*}
&  \frac{n_{1}}{n}\left[  g\left(  n^{\prime}\right)  -g\left(  n\right)
\right] \\
&  =\frac{n_{1}}{n}\left\{  \frac{1}{n+1}\left[  \sum_{s\neq1}f\left(
n_{s}\right)  +f\left(  n_{1}+1\right)  -f\left(  n+1\right)  \right]
-\frac{1}{n}\left[  \sum_{s}f\left(  n_{s}\right)  -f\left(  n\right)
\right]  \right\} \\
&  =\frac{n_{1}}{n}\left\{  \frac{\sum_{s}f\left(  n_{s}\right)  }{n+1}%
-\frac{f\left(  n+1\right)  }{n+1}+\frac{f\left(  n_{1}+1\right)  -f\left(
n_{1}\right)  }{n+1}+\frac{f\left(  n\right)  }{n}-\frac{\sum_{s}f\left(
n_{s}\right)  }{n}\right\} \\
&  =\frac{n_{1}}{n}\left\{  -\frac{f\left(  n+1\right)  }{n+1}-\frac{\sum
_{s}f\left(  n_{s}\right)  }{n\left(  n+1\right)  }+\frac{f\left(
n_{1}+1\right)  -f\left(  n_{1}\right)  }{n+1}+\frac{f\left(  n\right)  }%
{n}\right\} \\
&  =\frac{n_{1}}{n}\left\{  -\frac{f\left(  n+1\right)  }{n+1}-\frac{ng\left(
n\right)  +f\left(  n\right)  }{n\left(  n+1\right)  }+\frac{f\left(
n_{1}+1\right)  -f\left(  n_{1}\right)  }{n+1}+\frac{f\left(  n\right)  }%
{n}\right\} \\
&  =\frac{n_{1}}{n\left(  n+1\right)  }\left\{  -\delta_{1}\left(  n\right)
+\delta_{1}\left(  n_{1}\right)  -g\left(  n\right)  \right\} \\
&  =\frac{1}{n\left(  n+1\right)  }\left\{  n_{1}\delta_{1}\left(
n_{1}\right)  -\frac{n_{1}}{n}\cdot n\delta_{1}\left(  n\right)  -\frac{n_{1}%
}{n}\left[  \sum_{s}f\left(  n_{s}\right)  -f\left(  n\right)  \right]
\right\}
\end{align*}
From the previous Lemma,%
\[
0<x\delta_{1}\left(  x\right)  <\frac{1}{2}\text{, }0<f\left(  x\right)
<\frac{1}{2}\text{,}%
\]
so%
\[
-\frac{S}{2n^{2}}<-\frac{n_{1}}{n}\frac{S}{2n^{2}}<\frac{n_{1}}{n}\left[
g\left(  n^{\prime}\right)  -g\left(  n\right)  \right]  <\frac{1}{2n^{2}%
}\text{,}%
\]
thus%
\[
\frac{n_{1}}{n}\left\vert g\left(  n^{\prime}\right)  -g\left(  n\right)
\right\vert <\frac{S}{2n^{2}}\text{.}%
\]

\end{proof}%

%TCIMACRO{\TeXButton{endshaded}{\end{shaded}}}%
%BeginExpansion
\end{shaded}%
%EndExpansion

\subsection{Bounding the Difference Between $q_{\alpha}$ and $\tilde
{q}_{\alpha}$}

In this subsection we present the result bounding the difference between
$q_{\alpha}$ and $\tilde{q}_{\alpha}$, without making any assumptions to the
environment. Let $c_{\alpha}=\min_{s}\min_{a}\alpha_{s,a}$, the main
conclusion of this subsection is that%
\[
\left\vert q_{\alpha}\left(  s,a\right)  -\tilde{q}_{\alpha}\left(
s,a\right)  \right\vert \sim\frac{1}{c_{\alpha}^{2}}\text{.}%
\]

\begin{lemma}
\label{lem-04}$q_{\alpha}\left(  s,a\right)  \leq\frac{S-1}{2\left(
1-\gamma\right)  c_{\alpha}}$.
\end{lemma}

%

%TCIMACRO{\TeXButton{beginshaded}{\begin{shaded}}}%
%BeginExpansion
\begin{shaded}%
%EndExpansion

\begin{proof}
From Lemma.\ref{lem-g2}, write $K_{0}=\frac{S-1}{2}$, then%
\[
g\left(  \alpha_{s,a}\right)  <\frac{K_{0}}{\alpha_{s,a}}<\frac{K_{0}%
}{c_{\alpha}}\text{, }\forall s,a
\]
By definition,%
\begin{align*}
q_{\alpha}^{\gamma,2}\left(  s,a\right)   &  =g\left(  \alpha_{s,a}\right)
+\gamma\sum_{s^{\prime}}\frac{\alpha_{s,a,s^{\prime}}}{\alpha_{s,a}}%
\max_{a^{\prime}}q_{\alpha\triangleleft\left\langle s,a,s^{\prime
}\right\rangle }^{\gamma,1}\left(  s^{\prime},a^{\prime}\right) \\
&  <K_{0}\left(  \frac{1}{c_{\alpha}}+\gamma\frac{1}{c_{\alpha\triangleleft
\left\langle s,a,s^{\prime}\right\rangle }}\right) \\
&  \leq\frac{K_{0}}{c_{\alpha}}\left(  1+\gamma\right)  \text{,}%
\end{align*}
since $c_{\alpha\triangleleft\left\langle s,a,s^{\prime}\right\rangle }\geq
c_{\alpha}$. Repeat the process, it follows that for any $\tau$,%
\[
q_{\alpha}^{\gamma,\tau}\left(  s,a\right)  \leq\frac{K_{0}}{c_{\alpha}%
}\left(  1+\gamma+\cdots+\gamma^{\tau-1}\right)  <\frac{K_{0}}{\left(
1-\gamma\right)  c_{\alpha}}=\frac{S-1}{2\left(  1-\gamma\right)  c_{\alpha}%
}\text{,}%
\]
thus
\[
q_{\alpha}^{\gamma}\left(  s,a\right)  =\lim_{\tau\rightarrow\infty}q_{\alpha
}^{\gamma,\tau}\left(  s,a\right)  \leq\frac{K_{0}}{\left(  1-\gamma\right)
c_{\alpha}}.
\]

\end{proof}%

%TCIMACRO{\TeXButton{endshaded}{\end{shaded}}}%
%BeginExpansion
\end{shaded}%
%EndExpansion

\begin{lemma}
\label{lem-05}Let $n=\left[  n_{1},\cdots,n_{S}\right]  $, and $n^{\prime
}=\left[  n_{1}+1,n_{2},\cdots,n_{S}\right]  $. Let $x_{1},\cdots,x_{S}$ be
$S$ non-negative numbers. Define $p_{s}=\frac{n_{s}}{n}$, $p_{1}^{\prime
}=\frac{n_{1}+1}{n+1}$ and $p_{s}^{\prime}=\frac{n_{s}}{n+1}$ for
$s=2,\cdots,S$. Then%
\[
p_{1}\left\vert \sum_{s}\left(  p_{s}^{\prime}-p_{s}\right)  x_{s}\right\vert
\leq\frac{1}{n}\sum_{s}p_{s}x_{s}\text{.}%
\]

\end{lemma}%

%TCIMACRO{\TeXButton{beginshaded}{\begin{shaded}}}%
%BeginExpansion
\begin{shaded}%
%EndExpansion

\begin{proof}
Simple derivation gives%
\begin{align*}
p_{1}\left\vert \sum_{s}\left(  p_{s}^{\prime}-p_{s}\right)  x_{s}\right\vert
&  =\left(  \sum_{s}p_{s}x_{s}\right)  \cdot p_{1}\left\vert \frac{\sum
_{s}\left(  p_{s}^{\prime}-p_{s}\right)  x_{s}}{\sum_{s}p_{s}x_{s}}\right\vert
\\
&  \leq\left(  \sum_{s}p_{s}x_{s}\right)  \max_{s}\frac{p_{1}\cdot\left\vert
p_{s}^{\prime}-p_{s}\right\vert }{p_{s}}\text{.}%
\end{align*}
If $s=1$,%
\[
\frac{p_{1}\cdot\left\vert p_{1}^{\prime}-p_{1}\right\vert }{p_{1}}%
=\frac{n_{1}}{n}\cdot\frac{\frac{n_{1}+1}{n+1}-\frac{n_{1}}{n}}{\frac{n_{1}%
}{n}}=\frac{n-n_{1}}{n\left(  n+1\right)  }\leq\frac{1}{n}\text{.}%
\]
If $s\neq1$,%
\[
\frac{p_{1}\cdot\left\vert p_{s}^{\prime}-p_{s}\right\vert }{p_{s}}%
=\frac{n_{1}}{n}\cdot\frac{\frac{n_{s}}{n}-\frac{n_{s}}{n+1}}{\frac{n_{s}}{n}%
}=\frac{n_{1}}{n\left(  n+1\right)  }\leq\frac{1}{n}\text{.}%
\]
Therefore,%
\[
\max_{s}\frac{p_{1}\cdot\left\vert p_{s}^{\prime}-p_{s}\right\vert }{p_{s}%
}\leq\frac{1}{n}\text{,}%
\]
and%
\[
p_{1}\left\vert \sum_{s}\left(  p_{s}^{\prime}-p_{s}\right)  x_{s}\right\vert
\leq\frac{1}{n}\sum_{s}p_{s}x_{s}\text{.}%
\]

\end{proof}%

%TCIMACRO{\TeXButton{endshaded}{\end{shaded}}}%
%BeginExpansion
\end{shaded}%
%EndExpansion

\begin{lemma}
\label{lem-06}For any $\alpha$, $s^{\dag}$, $a^{\dag}$, there is some constant
$K$ depending on $S$ and $\gamma$ only, such that%
\[
\sum_{s}\frac{\alpha_{s^{\dag},a^{\dag},s}}{\alpha_{s^{\dag},a^{\dag}}}%
\max_{a}\left\vert q_{\alpha\triangleleft\left\langle s^{\dag},a^{\dag
},s\right\rangle }\left(  s,a\right)  -q_{\alpha}\left(  s,a\right)
\right\vert \leq\frac{K}{c_{\alpha}^{2}}\text{.}%
\]

\end{lemma}%

%TCIMACRO{\TeXButton{beginshaded}{\begin{shaded}}}%
%BeginExpansion
\begin{shaded}%
%EndExpansion

\begin{proof}
First change the notations. Let $s_{0}=s^{\dag}$, $a_{0}=a^{\dag}$. Also let
$\alpha^{1}=\alpha$, $\beta^{1}=\alpha\triangleleft\left\langle s^{\dag
},a^{\dag},s\right\rangle $. The result to prove becomes%
\[
\sum_{s_{1}}\frac{\alpha_{s_{0}a_{0}s_{1}}^{1}}{\alpha_{s_{0}a_{0}}^{1}}%
\max_{a_{1}}\left\vert q_{\beta^{1}}\left(  s_{1},a_{1}\right)  -q_{\alpha
^{1}}\left(  s_{1},a_{1}\right)  \right\vert \leq\frac{K}{c_{\alpha}^{2}%
}\text{.}%
\]

Consider the finite time horizon approximations of $q_{\beta^{1}}$ and
$q_{\alpha^{1}}$, namely $q_{\beta^{1}}^{\gamma,\tau}$ and $q_{\alpha^{1}%
}^{\gamma,\tau}$. With a little abuse of notation, we drop the superscript
$\gamma$ in this proof. Note that this shall not be confused with the finite
time horizon curiosity Q-values without discount.

For $\tau=2$, consider the following inequality:%
\begin{align*}
&  \frac{\alpha_{s_{0}a_{0}s_{1}}^{1}}{\alpha_{s_{0}a_{0}}^{1}}\max_{a_{1}%
}\left\vert q_{\beta^{1}}^{2}\left(  s_{1},a_{1}\right)  -q_{\alpha^{1}}%
^{2}\left(  s_{1},a_{1}\right)  \right\vert \\
&  \leq\frac{\alpha_{s_{0}a_{0}s_{1}}^{1}}{\alpha_{s_{0}a_{0}}^{1}}\max
_{a_{1}}\left\vert g\left(  \beta_{s_{1},a_{1}}^{1}\right)  -g\left(
\alpha_{s_{1},a_{1}}^{1}\right)  \right\vert \\
&  +\gamma\frac{\alpha_{s_{0}a_{0}s_{1}}^{1}}{\alpha_{s_{0}a_{0}}^{1}}%
\max_{a_{1}}\left\vert \sum_{s_{2}}\left(  \frac{\beta_{s_{1}a_{1}s_{2}}^{1}%
}{\beta_{s_{1}a_{1}}^{1}}-\frac{\alpha_{s_{1}a_{1}s_{2}}^{1}}{\alpha
_{s_{1}a_{1}}^{1}}\right)  \max_{a_{2}}q_{\beta^{2}}^{1}\left(  s_{2}%
,a_{2}\right)  \right\vert \\
&  +\gamma\frac{\alpha_{s_{0}a_{0}s_{1}}^{1}}{\alpha_{s_{0}a_{0}}^{1}}%
\max_{a_{1}}\sum_{s_{2}}\frac{\alpha_{s_{1}a_{1}s_{2}}^{1}}{\alpha_{s_{1}%
a_{1}}^{1}}\max_{a_{2}}\left\vert q_{\beta^{2}}^{1}\left(  s_{2},a_{2}\right)
-q_{\alpha^{2}}^{1}\left(  s_{2},a_{2}\right)  \right\vert \text{.}%
\end{align*}
Here $\beta^{2}=\beta^{1}\triangleleft\left\langle s_{1}a_{1}s_{2}%
\right\rangle $, $\alpha^{2}=\alpha^{1}\triangleleft\left\langle s_{1}%
a_{1}s_{2}\right\rangle $. Note that the error between $q_{\beta^{1}}^{2}$ and
$q_{\alpha^{1}}^{2}$ has been decomposed into three terms. The first term
captures the difference between the immediate information gain, the second
term captures the error between transition probabilities, and the third term
is of the same form as the left side of the inequality, except $\tau$ is
decreased by $1$. To simplify the notation, let $\mathbb{F}^{t}$ be the
operator%
\[
\mathbb{F}^{t}\left[  \cdots\right]  =\sum_{s_{t}}\frac{\alpha_{s_{t-1}%
a_{t-1}s_{t}}^{t-1}}{\alpha_{s_{t-1}a_{t-1}}^{t-1}}\max_{a_{t}}\left[
\cdots\right]  \text{.}%
\]
For fixed $\tau$, let%
\begin{align*}
\delta_{t}  &  =\left\vert g\left(  \beta_{s_{t},a_{t}}^{t}\right)  -g\left(
\alpha_{s_{t},a_{t}}^{t}\right)  \right\vert \\
&  +\gamma\left\vert \sum_{s_{t+1}}\left(  \frac{\beta_{s_{t}a_{t}s_{t+1}}%
^{t}}{\beta_{s_{t}a_{t}}^{t}}-\frac{\alpha_{s_{t}a_{t}s_{t+1}}^{t}}%
{\alpha_{s_{t}a_{t}}^{t}}\right)  \max_{a_{t+1}}q_{\beta^{t+1}}^{\tau
-t}\left(  s_{t+1},a_{t+1}\right)  \right\vert \text{,}%
\end{align*}
and%
\[
\phi_{t}=q_{\beta^{t}}^{\tau+1-t}\left(  s_{t},a_{t}\right)  -q_{\alpha^{t}%
}^{\tau+1-t}\left(  s_{t},a_{t}\right)  \text{.}%
\]
One can write%
\[
\mathbb{F}^{1}\phi_{1}\leq\mathbb{F}^{1}\delta_{1}+\gamma\mathbb{F}%
^{1}\mathbb{F}^{2}\phi_{2}\text{.}%
\]
Repeat this process for general $\tau$, it follows that%
\begin{align*}
\mathbb{F}^{1}\phi_{1}  &  \leq\mathbb{F}^{1}\delta_{1}+\gamma\mathbb{F}%
^{1}\mathbb{F}^{2}\phi_{2}\\
&  \leq\mathbb{F}^{1}\delta_{1}+\gamma\mathbb{F}^{1}\left[  \mathbb{F}%
^{2}\delta_{2}+\gamma\mathbb{F}^{2}\mathbb{F}^{3}\phi_{3}\right] \\
&  =\mathbb{F}^{1}\delta_{1}+\gamma\mathbb{F}^{1}\mathbb{F}^{2}\delta
_{2}+\gamma^{2}\mathbb{F}^{1}\mathbb{F}^{2}\mathbb{F}^{3}\phi_{3}\\
&  =\cdots\\
&  =\mathbb{F}^{1}\delta_{1}+\gamma\mathbb{F}^{1}\mathbb{F}^{2}\delta
_{2}+\cdots+\gamma^{t-1}\mathbb{F}^{1}\cdots\mathbb{F}^{t}\delta_{t}%
+\cdots+\gamma^{\tau-2}\mathbb{F}^{1}\cdots\mathbb{F}^{\tau-1}\delta_{\tau
-1}\\
&  +\gamma^{\tau-1}\mathbb{F}^{1}\cdots\mathbb{F}^{\tau}\phi_{\tau}\text{.}%
\end{align*}

Now look at a particular term in the inequality above, for example,%
\[
\mathbb{F}^{1}\cdots\mathbb{F}^{t}\delta_{t}=\sum_{s_{1}}\frac{\alpha
_{s_{0}a_{0}s_{1}}^{1}}{\alpha_{s_{0}a_{0}}^{1}}\max_{a_{1}}\cdots\sum_{s_{t}%
}\frac{\alpha_{s_{t-1}a_{t-1}s_{t}}^{t-1}}{\alpha_{s_{t-1}a_{t-1}}^{t-1}}%
\max_{a_{t}}\delta_{t}\text{.}%
\]
Note that if $\left\langle s_{t},a_{t}\right\rangle \neq\left\langle s^{\dag
},a^{\dag}\right\rangle $ then $\delta_{t}=0$, since $\beta^{t}$ and
$\alpha^{t}$ differ only in the entry indexed by $\left\langle s^{\dag
},a^{\dag},s_{1}\right\rangle $. The following discussion assumes that
$\left\langle s_{t},a_{t}\right\rangle =\left\langle s^{\dag},a^{\dag
}\right\rangle $. From Lemma.\ref{lem-g3}, let $K_{1}=\frac{S}{2}$, then%
\[
\left\vert g\left(  \beta_{s_{t},a_{t}}^{t}\right)  -g\left(  \alpha
_{s_{t},a_{t}}^{t}\right)  \right\vert \leq\frac{K_{1}}{\alpha_{s_{t},a_{t}%
}^{t}\alpha_{s_{t},a_{t},s_{1}}^{t}}\text{.}%
\]
From Lemma.\ref{lem-04} and \ref{lem-05}, there is some $K_{2}$ depends only
on $S$ and $\gamma$, such that%
\begin{align*}
&  \left\vert \sum_{s_{t+1}}\left(  \frac{\beta_{s_{t}a_{t}s_{t+1}}^{t}}%
{\beta_{s_{t}a_{t}}^{t}}-\frac{\alpha_{s_{t}a_{t}s_{t+1}}^{t}}{\alpha
_{s_{t}a_{t}}^{t}}\right)  \max_{a_{t+1}}q_{\beta^{t+1}}^{\tau-t}\left(
s_{t+1},a_{t+1}\right)  \right\vert \\
&  \leq\frac{1}{\alpha_{s_{0},a_{0},s_{1}}^{t}}\sum_{s_{t+1}}\frac
{\alpha_{s_{t}a_{t}s_{t+1}}^{t}}{\alpha_{s_{t}a_{t}}^{t}}\max_{a_{t+1}%
}q_{\beta^{t+1}}^{\tau-t}\left(  s_{t+1},a_{t+1}\right) \\
&  \leq\frac{K_{2}}{\alpha_{s_{0},a_{0},s_{1}}^{t}c_{\alpha^{t}}}\text{,}%
\end{align*}
where $c_{\alpha^{t}}=\min_{s}\min_{a}\alpha_{s,a}^{t}$. In combination, there
is some $K_{0}$ such that%
\[
\delta_{t}\leq\frac{K_{0}}{c_{\alpha^{t}}\alpha_{s^{\dag},a^{\dag},s_{1}}^{t}%
}\text{.}%
\]

The next step is tricky: Assume that the policy is given, say, it is already
the policy maximize $\mathbb{F}^{1}\cdots\mathbb{F}^{t}\delta_{t}$, so that
each $a$ is a deterministic function of the prior $\alpha^{1}$ and the
previous history. Consider a trajectory $s_{0}a_{0}s_{1}a_{1}\cdots s_{t}%
a_{t}$, the predictive probability of seeing such a trajectory is given by%
\[
p\left(  s_{1}a_{1}\cdots s_{t}a_{t}|s_{0}a_{0}\right)  =\frac{\alpha
_{s_{0}a_{0}s_{1}}^{1}}{\alpha_{s_{0}a_{0}}^{1}}\frac{\alpha_{s_{1}a_{1}s_{2}%
}^{2}}{\alpha_{s_{1}a_{1}}^{2}}\cdots\frac{\alpha_{s_{t-1}a_{t-1}s_{t}}^{t-1}%
}{\alpha_{s_{t-1}a_{t-1}}^{t-1}}\text{.}%
\]
Again, if $\left\langle s_{t},a_{t}\right\rangle \neq\left\langle s^{\dag
},a^{\dag}\right\rangle $, then $p\left(  s_{1}a_{1}\cdots s_{t}a_{t}%
|s_{0}a_{0}\right)  \delta_{t}=0$. Otherwise,%
\begin{align*}
p\left(  s_{1}a_{1}\cdots s_{t}a_{t}|s_{0}a_{0}\right)  \delta_{t}  &
=\frac{\alpha_{s_{1}a_{1}s_{2}}^{2}}{\alpha_{s_{1}a_{1}}^{2}}\cdots
\frac{\alpha_{s_{t-1}a_{t-1}s_{t}}^{t-1}}{\alpha_{s_{t-1}a_{t-1}}^{t-1}}%
\cdot\left(  \frac{\alpha_{s_{0}a_{0}s_{1}}^{1}}{\alpha_{s_{0}a_{0}}^{1}%
}\delta_{t}\right) \\
&  \leq\frac{\alpha_{s_{1}a_{1}s_{2}}^{2}}{\alpha_{s_{1}a_{1}}^{2}}\cdots
\frac{\alpha_{s_{t-1}a_{t-1}s_{t}}^{t-1}}{\alpha_{s_{t-1}a_{t-1}}^{t-1}}%
\cdot\frac{K_{0}}{c_{\alpha^{t}}\alpha_{s_{0}a_{0}}^{1}}\cdot\frac
{\alpha_{s_{0}a_{0}s_{1}}^{1}}{\alpha_{s_{0},a_{0},s_{1}}^{t}}\\
&  \leq\frac{\alpha_{s_{1}a_{1}s_{2}}^{2}}{\alpha_{s_{1}a_{1}}^{2}}\cdots
\frac{\alpha_{s_{t-1}a_{t-1}s_{t}}^{t-1}}{\alpha_{s_{t-1}a_{t-1}}^{t-1}}%
\cdot\frac{K_{0}}{c_{\alpha^{t}}\alpha_{s_{0}a_{0}}^{1}}\\
&  \leq\frac{\alpha_{s_{1}a_{1}s_{2}}^{2}}{\alpha_{s_{1}a_{1}}^{2}}\cdots
\frac{\alpha_{s_{t-1}a_{t-1}s_{t}}^{t-1}}{\alpha_{s_{t-1}a_{t-1}}^{t-1}}%
\cdot\frac{K_{0}}{c_{\alpha^{1}}^{2}}\text{.}%
\end{align*}
Note that%
\[
\frac{\alpha_{s_{1}a_{1}s_{2}}^{2}}{\alpha_{s_{1}a_{1}}^{2}}\cdots\frac
{\alpha_{s_{t-1}a_{t-1}s_{t}}^{t-1}}{\alpha_{s_{t-1}a_{t-1}}^{t-1}}=p\left(
s_{2}a_{2}\cdots s_{t}a_{t}|s_{1}a_{1}\right)
\]
is the probability of seeing the trajectory $s_{2}a_{2}\cdots s_{t}a_{t}$,
when the agent assumes prior $\alpha^{1}\triangleleft\left\langle s_{0}%
,a_{0},s_{1}\right\rangle =\alpha^{2}$, and follows the same policy starting
from $\left\langle s_{1},a_{1}\right\rangle $. Clearly,%
\[
\sum_{s_{2}}\cdots\sum_{s_{t}}p\left(  s_{2}a_{2}\cdots s_{t}a_{t}|s_{1}%
a_{1}\right)  =1\text{,}%
\]
which leads to%
\begin{align*}
\mathbb{F}^{1}\cdots\mathbb{F}^{t}\delta_{t}  &  =\sum_{s_{1}}\sum_{s_{2}%
}\cdots\sum_{s_{t}}p\left(  s_{1}a_{1}\cdots s_{t}a_{t}|s_{0}a_{0}\right)
\delta_{t}\\
&  \leq\sum_{s_{1}}\frac{K_{0}}{c_{\alpha^{1}}^{2}}\sum_{s_{2}}\cdots
\sum_{s_{t}}p\left(  s_{2}a_{2}\cdots s_{t}a_{t}|s_{1}a_{1}\right) \\
&  \leq\frac{SK_{0}}{c_{\alpha^{1}}^{2}}\text{.}%
\end{align*}

Putting the equation back, and note that $c_{\alpha^{1}}^{2}=c_{\alpha}^{2}$
is a constant on $\alpha$, one has%
\begin{align*}
\mathbb{F}^{1}\phi_{1}  &  \leq\mathbb{F}^{1}\delta_{1}+\gamma\mathbb{F}%
^{1}\mathbb{F}^{2}\delta_{2}+\cdots+\gamma^{t-1}\mathbb{F}^{1}\cdots
\mathbb{F}^{t}\delta_{t}+\cdots+\gamma^{\tau-2}\mathbb{F}^{1}\cdots
\mathbb{F}^{\tau-1}\delta_{\tau-1}\\
&  +\gamma^{\tau-1}\mathbb{F}^{1}\cdots\mathbb{F}^{\tau}\phi_{\tau}\\
&  \leq\frac{SK_{0}}{c_{\alpha}^{2}}\left(  1+\gamma+\cdots+\gamma^{\tau
-2}\right)  +\gamma^{\tau-1}\mathbb{F}^{1}\cdots\mathbb{F}^{\tau}\phi_{\tau}\\
&  \leq\frac{SK_{0}}{1-\gamma}\frac{1}{c_{\alpha}}+\gamma^{\tau-1}%
\mathbb{F}^{1}\cdots\mathbb{F}^{\tau}\phi_{\tau}\text{.}%
\end{align*}
From Lemma.\ref{lem-04}, since the curiosity Q-values are bounded, there is
some $K_{3}$ such that%
\begin{align*}
\phi_{\tau}  &  =\left\vert q_{\beta^{t}}^{1}\left(  s_{t},a_{t}\right)
-q_{\alpha^{t}}^{1}\left(  s_{t},a_{t}\right)  \right\vert \\
&  \leq\left\vert q_{\beta^{t}}^{1}\left(  s_{t},a_{t}\right)  \right\vert
+\left\vert q_{\alpha^{t}}^{1}\left(  s_{t},a_{t}\right)  \right\vert \\
&  \leq\left\vert q_{\beta^{t}}\left(  s_{t},a_{t}\right)  \right\vert
+\left\vert q_{\alpha^{t}}\left(  s_{t},a_{t}\right)  \right\vert \\
&  \leq\frac{K_{3}}{c_{\alpha}}\text{,}%
\end{align*}
thus%
\[
\mathbb{F}^{1}\phi_{1}\leq\frac{SK_{0}}{1-\gamma}\frac{1}{c_{\alpha}}%
+\gamma^{\tau-1}\frac{K_{3}}{c_{\alpha}}\text{.}%
\]
Let $\tau\rightarrow\infty$, one has%
\[
\sum_{s_{1}}\frac{\alpha_{s_{0}a_{0}s_{1}}^{1}}{\alpha_{s_{0}a_{0}}^{1}}%
\max_{a_{1}}\left\vert q_{\beta^{1}}\left(  s_{1},a_{1}\right)  -q_{\alpha
^{1}}\left(  s_{1},a_{1}\right)  \right\vert \leq\frac{K}{c_{\alpha}^{2}%
}\text{,}%
\]
where $K=\frac{SK_{0}}{1-\gamma}$ is a constant depending on $S$ and $\gamma$ only.
\end{proof}

%

%TCIMACRO{\TeXButton{endshaded}{\end{shaded}}}%
%BeginExpansion
\end{shaded}%
%EndExpansion

The central result in this subsection is given by the following Proposition.

\begin{proposition}
\label{prop-04}There is some $K>0$ depending on $S$ and $\gamma$ only, such
that%
\[
\left\vert q_{\alpha}\left(  s,a\right)  -\tilde{q}_{\alpha}\left(
s,a\right)  \right\vert \leq\frac{K}{c_{\alpha}^{2}}\text{.}%
\]
\end{proposition}%

%TCIMACRO{\TeXButton{beginshaded}{\begin{shaded}}}%
%BeginExpansion
\begin{shaded}%
%EndExpansion

\begin{proof}
Write Eq.\ref{eq-05} into the following form%
\begin{align*}
q_{\alpha}\left(  s,a\right)   &  =g\left(  \alpha_{s,a}\right)  +\gamma
\sum_{s^{\prime}}\frac{\alpha_{s,a,s^{\prime}}}{\alpha_{s,a}}\max_{a^{\prime}%
}q_{\alpha}\left(  s^{\prime},a^{\prime}\right) \\
&  +\gamma\sum_{s^{\prime}}\frac{\alpha_{s,a,s^{\prime}}}{\alpha_{s,a}}\left[
\max_{a^{\prime}}q_{\alpha\triangleleft\left\langle s,a,s^{\prime
}\right\rangle }\left(  s^{\prime},a^{\prime}\right)  -\max_{a^{\prime}%
}q_{\alpha}\left(  s^{\prime},a^{\prime}\right)  \right]  \text{.}%
\end{align*}
The last term is bounded by%
\begin{align*}
\delta &  =\sum_{s^{\prime}}\frac{\alpha_{s,a,s^{\prime}}}{\alpha_{s,a}%
}\left\vert \max_{a^{\prime}}q_{\alpha\triangleleft\left\langle s,a,s^{\prime
}\right\rangle }\left(  s^{\prime},a^{\prime}\right)  -\max_{a^{\prime}%
}q_{\alpha}\left(  s^{\prime},a^{\prime}\right)  \right\vert \\
&  \leq\sum_{s^{\prime}}\frac{\alpha_{s,a,s^{\prime}}}{\alpha_{s,a}}%
\max_{a^{\prime}}\left\vert q_{\alpha\triangleleft\left\langle s,a,s^{\prime
}\right\rangle }\left(  s^{\prime},a^{\prime}\right)  -q_{\alpha}\left(
s^{\prime},a^{\prime}\right)  \right\vert \text{.}%
\end{align*}
Apply Lemma.\ref{lem-05}, it follows that there is some constant $K_{0}$
depending on $S$ and $\gamma$ only, such that%
\[
\sum_{s^{\prime}}\frac{\alpha_{s,a,s^{\prime}}}{\alpha_{s,a}}\max_{a^{\prime}%
}\left\vert q_{\alpha\triangleleft\left\langle s,a,s^{\prime}\right\rangle
}\left(  s^{\prime},a^{\prime}\right)  -q_{\alpha}\left(  s^{\prime}%
,a^{\prime}\right)  \right\vert \leq\frac{K_{0}}{c_{\alpha}^{2}}\text{.}%
\]
Therefore $\delta\leq\frac{K_{0}}{c_{\alpha}^{2}}$.

Now compare $q_{\alpha}$ and $\tilde{q}_{\alpha}$:%
\begin{align*}
&  \max_{s}\max_{a}\left\vert q_{\alpha}\left(  s,a\right)  -\tilde{q}%
_{\alpha}\left(  s,a\right)  \right\vert \\
&  =\max_{s}\max_{a}\left\vert \gamma\sum_{s^{\prime}}\frac{\alpha
_{s,a,s^{\prime}}}{\alpha_{s,a}}\left(  \max_{a^{\prime}}q_{\alpha}\left(
s^{\prime},a^{\prime}\right)  -\max_{s^{\prime}}\tilde{q}_{\alpha}\left(
s^{\prime},a^{\prime}\right)  \right)  +\gamma\delta\right\vert \\
&  \leq\gamma\max_{s}\max_{a}\left\vert q_{\alpha}\left(  s,a\right)
-\tilde{q}_{\alpha}\left(  s,a\right)  \right\vert +\frac{\gamma K_{0}%
}{c_{\alpha}^{2}}\text{.}%
\end{align*}
Therefore%
\[
\max_{s}\max_{a}\left\vert q_{\alpha}\left(  s,a\right)  -\tilde{q}_{\alpha
}\left(  s,a\right)  \right\vert \leq\frac{\gamma K_{0}}{1-\gamma}\cdot
\frac{1}{c_{\alpha}^{2}}\text{.}%
\]
Letting $K=\frac{\gamma K_{0}}{1-\gamma}$ completes the proof.
\end{proof}

%

%TCIMACRO{\TeXButton{endshaded}{\end{shaded}}}%
%BeginExpansion
\end{shaded}%
%EndExpansion

\subsection{Quality of the Approximation in Connected Markovian Environment}

Proposition.\ref{prop-04} guarantees that the difference between $q_{\alpha}$
and $\tilde{q}_{\alpha}$ decreases at the rate of $c_{\alpha}^{-2}$. However,
this alone is not enough to guarantee that $\tilde{q}_{\alpha}$ converges to
$q_{\alpha}$ when the agent operates in the environment. For example, consider
the environment consists of two connected components. In this case,
$c_{\alpha}$ is upper bounded since that in one of the connected component
$\alpha_{s,a}$ never increases. Here we make the following assumption:

\begin{description}
\item[Assumption III] The environment is finite Markovian with dynamics
$p\left(  s^{\prime}|s,a\right)  $, and the Markov chain with transition
kernel%
\[
p\left(  s^{\prime}|s\right)  =\frac{1}{A}\sum_{a\in\mathcal{A}}p\left(
s^{\prime}|s,a\right)
\]
is irreducible.
\end{description}

The first half of the assumption ensures that $\frac{\alpha_{s,a,s^{\prime}}%
}{\alpha_{s,a}}$ converges to $p\left(  s^{\prime}|s,a\right)  $ when
$\alpha_{s,a}$ goes to infinity by Law of Large Numbers. The second half
of\ the assumption implies that it is always possible to navigate from one
state to another with positive probability of success. Therefore, if some
$g\left(  \alpha_{s,a}\right)  $ is large, the information is guaranteed to
propagate to all the states. Under this assumption, we prove in this section
that when $t\rightarrow\infty$,%
\[
\left\vert \frac{q_{\alpha}\left(  s,a\right)  }{\tilde{q}_{\alpha}\left(
s,a\right)  }-1\right\vert \rightarrow0\text{,}%
\]
namely, the curiosity Q-value and the DP approximation are getting arbitrarily
closer along time.

The proof is unwrapped in three steps.

\begin{lemma}
\label{lem-e1}Assume IV), and the agent chooses the action greedily with
respect to $\tilde{q}_{\alpha^{t}}$, where $\alpha^{t}$ is the posterior after
$t$ time steps. Then for any $s,a$,%
\[
\lim_{t\rightarrow\infty}\alpha_{s,a}^{t}=\infty\text{, a.s.}%
\]

\end{lemma}%

%TCIMACRO{\TeXButton{beginshaded}{\begin{shaded}}}%
%BeginExpansion
\begin{shaded}%
%EndExpansion

\begin{proof}
Note that $\alpha_{s,a,s^{\prime}}^{t}$ is non-decreasing, and can only
increase by one if increasing. Therefore, $\lim_{t\rightarrow\infty}%
\alpha_{s,a}^{t}<\infty$ implies that there is some $T_{s,a}$ and $c_{s,a}$
such that for all $t>T$, $\alpha_{s,a}^{t}=c_{s,a}$.

The complement of $\lim_{t\rightarrow\infty}\alpha_{s,a}^{t}=\infty$ for all
$\left\langle s,a\right\rangle $ is that $\exists\Lambda\subset\mathcal{S}%
\times\mathcal{A}$, $\Lambda\neq\emptyset$, and $\exists T_{s,a},c_{s,a}$ for
all $\left\langle s,a\right\rangle \in\Lambda$, such that $\alpha_{s,a}%
^{t}=c_{s,a}$ for all $t>T_{s,a}$. Since there are only finitely many
$\left\langle s,a\right\rangle $, this can be simplified to $\exists
\Lambda\neq\emptyset$, $\exists T$, $\exists c_{s,a}$, such that $\alpha
_{s,a}^{t}=c_{s,a}$ for all $t>T$ and $\left\langle s,a\right\rangle
\in\Lambda$.

Fix $\Lambda\neq\emptyset$, $T$ and $c_{s,a}$, we show that the event
$\alpha_{s,a}^{t}=c_{s,a}$ for $t>T$ and $\left\langle s,a\right\rangle
\in\Lambda$ is a null event. Let $\bar{\Lambda}=\mathcal{S}\times
\mathcal{A}\backslash\Lambda$, by definition, $\alpha_{s,a}^{t}\rightarrow
\infty$ for all $\left\langle s,a\right\rangle \in\bar{\Lambda}$. Clearly,
$\bar{\Lambda}$ is not empty. Define%
\[
\mathcal{S}_{I}=\left\{  s\in\mathcal{S}:\exists a,a^{\prime\prime}\text{ such
that }\left\langle s,a\right\rangle \in\Lambda\text{, }\left\langle
s,a^{\prime\prime}\right\rangle \notin\Lambda\right\}  \text{.}%
\]
Namely, $\mathcal{S}_{I}$ is the `boundary' between $\Lambda$ and
$\bar{\Lambda}$.

The first step is to show $\mathcal{S}_{I}\neq\emptyset$ if $\Lambda
\neq\emptyset$, or more precisely, the event $\mathcal{S}_{I}=\emptyset$ and
$\Lambda\neq\emptyset$ is null. Assume $\mathcal{S}_{I}=\emptyset$ and
$\Lambda\neq\emptyset$, then $\Lambda$ must satisfy that if $\left\langle
s,a\right\rangle \in\Lambda$ for some $a$, then $\left\langle s,a\right\rangle
\in\Lambda$ for all $a$. Let $\mathcal{S}_{\Lambda}\subset\mathcal{S}$ be the
set of $s$ such that $\left\langle s,a\right\rangle \in\Lambda$. Clearly, once
reaching $s\in\mathcal{S}_{\Lambda}$, any action chosen would cause $\Lambda$
be visited, which can only happen for finitely many times. This implies that
for any $s\in\mathcal{S}_{\Lambda}$, any state action pair $\left\langle
s^{\prime},a^{\prime}\right\rangle $ such that $p\left(  s|s^{\prime
},a^{\prime}\right)  >0$ can only be visited finite number of times
\emph{almost surely}, because the probability of sampling from $p\left(
\cdot|s^{\prime},a^{\prime}\right)  $ for infinitely many times but only
getting finite number of $s$ is zero. From Assumption IV), for any
$\mathcal{S}_{\Lambda}\neq\mathcal{S}$, there is always some $\left\langle
s^{\prime},a^{\prime}\right\rangle $ such that $s^{\prime}\notin
\mathcal{S}_{\Lambda}$ and $p\left(  s|s^{\prime},a^{\prime}\right)  >0$, so
$\left\langle s^{\prime},a^{\prime}\right\rangle $ can only be visited
finitely many times, by definition $\left\langle s^{\prime},a^{\prime
}\right\rangle \in\Lambda$, which contradicts with the fact that $s^{\prime
}\notin\mathcal{S}_{\Lambda}$.

Next we show that at least for one $s\in\mathcal{S}_{I}$, following the
optimal strategy leads to some $\left\langle s,a\right\rangle \in\Lambda$
being visited. For $t>T$, Define%
\[
\hat{q}\left(  s,a\right)  =r\left(  s,a\right)  +\gamma\sum_{s^{\prime}}%
\hat{p}\left(  s^{\prime}|s,a\right)  \max_{a^{\prime}}\hat{q}\left(
s^{\prime},a^{\prime}\right)  \text{,}%
\]
with%
\[
\hat{p}\left(  s^{\prime}|s,a\right)  =\left\{
\begin{array}
[c]{l}%
p\left(  s^{\prime}|s,a\right)  \text{, if }\left\langle s,a\right\rangle
\notin\Lambda\\
\frac{\alpha_{s,a,s^{\prime}}^{t}}{\alpha_{s,a}^{t}}\text{, if }\left\langle
s,a\right\rangle \in\Lambda
\end{array}
\right.  \text{,}%
\]
and%
\[
r\left(  s,a\right)  =\left\{
\begin{array}
[c]{l}%
0\text{, if }\left\langle s,a\right\rangle \notin\Lambda\\
g\left(  \alpha_{s,a}^{t}\right)  \text{, if }\left\langle s,a\right\rangle
\in\Lambda
\end{array}
\right.  \text{.}%
\]
Clearly, $\hat{p}$ and $r$ do not depend on $t$, and $\hat{q}$ is the unique
optimal solution. Now let $\left\langle s^{\dag},a^{\dag}\right\rangle
\in\Lambda$ be the pair such that $s\in\mathcal{S}_{I}$, and%
\[
\hat{q}\left(  s^{\dag},a^{\dag}\right)  =\max_{\left\langle s,a\right\rangle
\in\Lambda,s\in\mathcal{S}_{I}}\hat{q}\left(  s,a\right)  \text{.}%
\]
It can be seen that for any $a^{\prime}$ such that $\left\langle s^{\dag
},a^{\prime}\right\rangle \notin\Lambda$, $\hat{q}\left(  s^{\dag},a^{\prime
}\right)  \leq\gamma\hat{q}\left(  s^{\dag},a^{\dag}\right)  $. The reason is
the following: Performing $a^{\prime}$ leads to zero immediate reward since
$\left\langle s^{\dag},a^{\prime}\right\rangle \notin\Lambda$. Let
$s^{\prime\prime}$ be the result of the transition, then either $s^{\prime
\prime}\in\mathcal{S}_{I}$, so $\max_{a^{\prime\prime}}\hat{q}\left(
s^{\prime\prime},a^{\prime\prime}\right)  \leq\hat{q}\left(  s^{\dag},a^{\dag
}\right)  $, or $s^{\prime\prime}$ is some other state such that $\left\langle
s^{\prime\prime},a^{\prime\prime}\right\rangle \notin\Lambda$ for all
$a^{\prime\prime}$. (Note that $s^{\prime\prime}$ cannot be a state such that
$\left\langle s^{\prime\prime},a^{\prime\prime}\right\rangle \in\Lambda$ for
all $a^{\prime\prime}$.) In the latter case, since $s^{\prime\prime}$ is only
connected to states in $\Lambda$ through $\mathcal{S}_{I}$, it must be that%
\[
\max_{a^{\prime\prime}}\hat{q}\left(  s^{\prime\prime},a^{\prime\prime
}\right)  \leq\gamma\hat{q}\left(  s^{\dag},a^{\dag}\right)  \text{,}%
\]
since at least one more step must be made to reach $\mathcal{S}_{I}$ first.
Taking into account the discount, it follows that%
\[
\hat{q}\left(  s^{\dag},a^{\dag}\right)  -\hat{q}\left(  s^{\dag},a^{\prime
}\right)  \geq\left(  1-\gamma\right)  \hat{q}\left(  s^{\dag},a^{\dag
}\right)  \text{.}%
\]
Replace $\hat{q}$ with $\tilde{q}_{\alpha^{t}}$ leads to%
\begin{align*}
\tilde{q}_{\alpha^{t}}\left(  s^{\dag},a^{\dag}\right)  -\tilde{q}_{\alpha
^{t}}\left(  s^{\dag},a^{\prime}\right)   &  \geq\left(  1-\gamma\right)
\hat{q}\left(  s^{\dag},a^{\dag}\right) \\
&  +\tilde{q}_{\alpha^{t}}\left(  s^{\dag},a^{\dag}\right)  -\hat{q}\left(
s^{\dag},a^{\dag}\right)  +\tilde{q}_{\alpha^{t}}\left(  s^{\dag},a^{\prime
}\right)  -\hat{q}\left(  s^{\dag},a^{\prime}\right)
\end{align*}
From the initial assumption, when $t>T$, $\left\langle s^{\dag},a^{\dag
}\right\rangle $ is never visited, also, the action is chosen greedily with
respect to $\tilde{q}_{\alpha^{t}}$. This implies that at least for one
$a^{\prime}$ such that $\left\langle s^{\dag},a^{\prime}\right\rangle
\notin\Lambda$,%
\[
\tilde{q}_{\alpha^{t}}\left(  s^{\dag},a^{\dag}\right)  -\tilde{q}_{\alpha
^{t}}\left(  s^{\dag},a^{\prime}\right)  \leq0\text{,}%
\]
or%
\begin{align*}
0  &  \geq\left(  1-\gamma\right)  \hat{q}\left(  s^{\dag},a^{\dag}\right)
+\tilde{q}_{\alpha^{t}}\left(  s^{\dag},a^{\dag}\right)  -\hat{q}\left(
s^{\dag},a^{\dag}\right)  +\tilde{q}_{\alpha^{t}}\left(  s^{\dag},a^{\prime
}\right)  -\hat{q}\left(  s^{\dag},a^{\prime}\right) \\
&  \geq\left(  1-\gamma\right)  \hat{q}\left(  s^{\dag},a^{\dag}\right)
-2\max_{s}\max_{a}\left\vert \tilde{q}_{\alpha^{t}}\left(  s,a\right)
-\hat{q}\left(  s,a\right)  \right\vert \text{,}%
\end{align*}
which leads to%
\[
\max_{s}\max_{a}\left\vert \tilde{q}_{\alpha^{t}}\left(  s,a\right)  -\hat
{q}\left(  s,a\right)  \right\vert \geq\frac{1-\gamma}{2}\hat{q}\left(
s^{\dag},a^{\dag}\right)  \text{.}%
\]
Note that%
\begin{align*}
\left\vert \tilde{q}_{\alpha^{t}}\left(  s,a\right)  -\hat{q}\left(
s,a\right)  \right\vert  &  \leq\left\vert g\left(  \alpha_{s,a}^{t}\right)
-r\left(  s,a\right)  \right\vert \\
&  +\gamma\sum_{s^{\prime}}\left\vert \frac{\alpha_{s,a,s^{\prime}}^{t}%
}{\alpha_{s,a}^{t}}-\hat{p}\left(  s^{\prime}|s,a\right)  \right\vert
\max_{a^{\prime}}\tilde{q}_{\alpha^{t}}\left(  s^{\prime},a^{\prime}\right) \\
&  +\gamma\sum_{s^{\prime}}\hat{p}\left(  s^{\prime}|s,a\right)
\max_{a^{\prime}}\left\vert \tilde{q}_{\alpha^{t}}\left(  s^{\prime}%
,a^{\prime}\right)  -\hat{q}\left(  s,a\right)  \right\vert \text{,}%
\end{align*}
so%
\begin{align*}
&  \max_{s}\max_{a}\left\vert \tilde{q}_{\alpha^{t}}\left(  s,a\right)
-\hat{q}\left(  s,a\right)  \right\vert \\
&  \leq\frac{1}{1-\gamma}\left\vert g\left(  \alpha_{s,a}^{t}\right)
-r\left(  s,a\right)  \right\vert \\
&  +\frac{\gamma}{1-\gamma}\max_{s}\max_{a}\sum_{s^{\prime}}\left\vert
\frac{\alpha_{s,a,s^{\prime}}^{t}}{\alpha_{s,a}^{t}}-\hat{p}\left(  s^{\prime
}|s,a\right)  \right\vert \max_{a^{\prime}}\tilde{q}_{\alpha^{t}}\left(
s^{\prime},a^{\prime}\right)  \text{.}%
\end{align*}
From Lemma.\ref{lem-g2},
\[
\left\vert g\left(  \alpha_{s,a}^{t}\right)  -r\left(  s,a\right)  \right\vert
=\max_{\left\langle s,a\right\rangle \notin\Lambda}g\left(  \alpha_{s,a}%
^{t}\right)  <\frac{S-1}{2\alpha_{s,a}^{t}}\rightarrow0\text{.}%
\]
Therefore, there is some $T^{\prime}$ such that $\left\vert g\left(
\alpha_{s,a}^{t}\right)  -r\left(  s,a\right)  \right\vert <\frac{1-\gamma}%
{4}\hat{q}\left(  s^{\dag},a^{\dag}\right)  $ for all $\left\langle
s,a\right\rangle \notin\Lambda$. Also note that%
\[
\tilde{q}_{\alpha^{t}}\left(  s^{\prime},a^{\prime}\right)  \leq\frac
{S-1}{2\left(  1-\gamma\right)  c_{\alpha}}\text{,}%
\]
where $c_{\alpha}=\min_{\left\langle s,a\right\rangle \in\Lambda}c_{s,a}$. Let
$K=\frac{\gamma\left(  S-1\right)  }{2\left(  1-\gamma\right)  ^{2}c_{\alpha}%
}$, then%
\begin{align*}
&  K\max_{s}\max_{a}\sum_{s^{\prime}}\left\vert \frac{\alpha_{s,a,s^{\prime}%
}^{t}}{\alpha_{s,a}^{t}}-\hat{p}\left(  s^{\prime}|s,a\right)  \right\vert
+\frac{1-\gamma}{4}\hat{q}\left(  s^{\dag},a^{\dag}\right) \\
&  \geq\frac{\gamma}{1-\gamma}\max_{s}\max_{a}\sum_{s^{\prime}}\left\vert
\frac{\alpha_{s,a,s^{\prime}}^{t}}{\alpha_{s,a}^{t}}-\hat{p}\left(  s^{\prime
}|s,a\right)  \right\vert \max_{a^{\prime}}\tilde{q}_{\alpha^{t}}\left(
s^{\prime},a^{\prime}\right) \\
&  +\frac{1}{1-\gamma}\left\vert g\left(  \alpha_{s,a}^{t}\right)  -r\left(
s,a\right)  \right\vert \\
&  \geq\max_{s}\max_{a}\left\vert \tilde{q}_{\alpha^{t}}\left(  s,a\right)
-\hat{q}\left(  s,a\right)  \right\vert \\
&  \geq\frac{1-\gamma}{2}\hat{q}\left(  s^{\dag},a^{\dag}\right)  \text{,}%
\end{align*}
thus%
\begin{align*}
\max_{\left\langle s,a\right\rangle \notin\Lambda}\sum_{s^{\prime}}\left\vert
\frac{\alpha_{s,a,s^{\prime}}^{t}}{\alpha_{s,a}^{t}}-p\left(  s^{\prime
}|s,a\right)  \right\vert  &  =\max_{s}\max_{a}\sum_{s^{\prime}}\left\vert
\frac{\alpha_{s,a,s^{\prime}}^{t}}{\alpha_{s,a}^{t}}-\hat{p}\left(  s^{\prime
}|s,a\right)  \right\vert \\
&  \geq\frac{1-\gamma}{4K}\hat{q}\left(  s^{\dag},a^{\dag}\right)  \text{,}%
\end{align*}
for all $t>T^{\prime}$. This implies that when $t\rightarrow\infty$, the
empirical ratio $\frac{\alpha_{s,a,s^{\prime}}^{t}}{\alpha_{s,a}^{t}}$ does
not converge to $p\left(  s^{\prime}|s,a\right)  $, which is a null event
because it contradicts the Strong Law of Large Numbers. This in turn implies
that for fixed $\Lambda$, $T$ and $c_{s,a}$, the event $\exists\Lambda
\neq\emptyset$, $\exists T$, $\exists c_{s,a}$, such that $\alpha_{s,a}%
^{t}=c_{s,a}$ for all $t>T$ and $\left\langle s,a\right\rangle \in\Lambda$ is null.

As the last step, notice that there are only countably many such events, and
since the union of countably many null events is still null, one can conclude
that $\lim_{t\rightarrow\infty}\alpha_{s,a}^{t}=\infty$ for all $\left\langle
s,a\right\rangle $ holds almost surely.
\end{proof}

%

%TCIMACRO{\TeXButton{endshaded}{\end{shaded}}}%
%BeginExpansion
\end{shaded}%
%EndExpansion

The next step is to show that all $\tilde{q}_{\alpha^{t}}\left(  s,a\right)  $
decreases at the rate lower bounded by $c_{\alpha}^{-1}$. Let $q_{s,a}$ be the
Q-value of performing $a$ at state $s$, assuming the reward is $1$ at all
states. Namely, $q_{s,a}$ is the solution of the following Bellman equation%
\[
q_{s,a}=1+\gamma\sum_{s^{\prime}}p\left(  s^{\prime}|s,a\right)
\sum_{a^{\prime}}q_{s^{\prime},a^{\prime}}\text{.}%
\]
Clearly, $q_{s,a}>0$ from Assumption IV). Define $q=\min_{s,a}q_{s,a}$. Also,
let%
\[
u_{s,a}=\frac{\sum_{s^{\prime}\neq s^{\ast}}f\left(  \alpha_{s,a,s^{\prime}%
}^{0}\right)  -f\left(  \sum_{s^{\prime}\neq s^{\ast}}\alpha_{s,a,s^{\prime}%
}^{0}\right)  }{2}\text{,}%
\]
where $\alpha^{0}$ is the initial $\alpha$ representing the agent's prior, and
$s^{\ast}=\arg\max_{s^{\prime}}\alpha_{s,a,s^{\prime}}^{0}$ as defined in
Lemma.\ref{lem-g2}. Define $u=\min_{s,a}u_{s,a}$.

\begin{lemma}
\label{lem-e2}Assume IV), and let $c_{\alpha^{t}}=\min_{s}\min_{a}\alpha
_{s,a}^{t}$, then%
\[
\lim\inf_{t\rightarrow\infty}c_{\alpha^{t}}\tilde{q}_{\alpha^{t}}\left(
s,a\right)  \geq uq\text{, a.s.}%
\]

\end{lemma}%

%TCIMACRO{\TeXButton{beginshaded}{\begin{shaded}}}%
%BeginExpansion
\begin{shaded}%
%EndExpansion

\begin{proof}
Let $\hat{q}_{\alpha^{t}}$ be the solution to the following Bellman equation:%
\[
\hat{q}_{\alpha^{t}}\left(  s,a\right)  =g\left(  \alpha_{s,a}^{t}\right)
+\gamma\sum_{s^{\prime}}p\left(  s^{\prime}|s,a\right)  \max_{a^{\prime}}%
\hat{q}_{\alpha^{t}}\left(  s^{\prime},a^{\prime}\right)  \text{.}%
\]
Clearly, for any $\left\langle s,a\right\rangle $,%
\[
g\left(  \alpha_{s,a}^{t}\right)  \geq\frac{u_{s,a}}{\alpha_{s,a}^{t}}%
\geq\frac{u}{c_{\alpha^{t}}}\text{.}%
\]
and because $\hat{q}_{\alpha^{t}}$ is optimal,
\[
\hat{q}_{\alpha^{t}}\left(  s,a\right)  \geq\frac{u}{c_{\alpha^{t}}}%
q_{s,a}\geq\frac{uq}{c_{\alpha^{t}}}\text{, }\forall s,a\text{,}%
\]
or $c_{\alpha^{t}}\hat{q}_{\alpha^{t}}\left(  s,a\right)  \geq uq$.

Fix an $\varepsilon>0$, we show that%
\[
\lim\inf_{t\rightarrow\infty}c_{\alpha^{t}}\tilde{q}_{\alpha^{t}}\left(
s,a\right)  \leq uq\left(  1-\varepsilon\right)  \text{, }\forall s,a
\]
is a null event. Assuming $\lim\inf_{t\rightarrow\infty}c_{\alpha^{t}}%
\tilde{q}_{\alpha^{t}}\leq uq\left(  1-\varepsilon\right)  $, and following
similar procedure as in the proof of Lemma.\ref{lem-e1}, let $K=\frac
{\gamma\left(  S-1\right)  }{2\left(  1-\gamma\right)  ^{2}}$, then%
\begin{align*}
\max_{s}\max_{a}\sum_{s^{\prime}}\left\vert \frac{\alpha_{s,a,s^{\prime}}^{t}%
}{\alpha_{s,a}^{t}}-\hat{p}\left(  s^{\prime}|s,a\right)  \right\vert  &
\geq\frac{c_{\alpha}}{K}\max_{s}\max_{a}\left\vert \tilde{q}_{\alpha^{t}%
}\left(  s,a\right)  -\hat{q}\left(  s,a\right)  \right\vert \\
&  \geq\frac{uq\varepsilon}{K}%
\end{align*}
holds for infinitely many $t$, which contradicts again with the Law of Large Numbers.

Let $\varepsilon_{n}=\frac{1}{n}$, then the union of the countably many events%
\[
\lim\inf_{t\rightarrow\infty}c_{\alpha^{t}}\tilde{q}_{\alpha^{t}}\left(
s,a\right)  \leq uq\left(  1-\varepsilon_{n}\right)  \text{, }\forall s,a
\]
is again a null event, therefore%
\[
\lim\inf_{t\rightarrow\infty}c_{\alpha^{t}}\tilde{q}_{\alpha^{t}}\left(
s,a\right)  \geq uq\text{, a.s.}%
\]

\end{proof}%

%TCIMACRO{\TeXButton{endshaded}{\end{shaded}}}%
%BeginExpansion
\end{shaded}%
%EndExpansion

Combining Lemma.\ref{lem-e1} and \ref{lem-e2} produces the following proposition.

\begin{proposition}
\label{prop-05}Assume IV), and that the agent acts greedily with respect to
$\tilde{q}_{\alpha}$, then%
\[
\lim_{t\rightarrow\infty}\left\vert \frac{q_{\alpha^{t}}}{\tilde{q}%
_{\alpha^{t}}}-1\right\vert =0\text{, a.s.}%
\]
and there is some $K$ depending only on the dynamics and $\gamma$, such that%
\[
\lim\sup_{t\rightarrow\infty}c_{\alpha}\left\vert \frac{q_{\alpha^{t}}}%
{\tilde{q}_{\alpha^{t}}}-1\right\vert \leq K\text{.}%
\]
\end{proposition}

%

%TCIMACRO{\TeXButton{beginshaded}{\begin{shaded}}}%
%BeginExpansion
\begin{shaded}%
%EndExpansion

\begin{proof}
Note that%
\[
\left\vert \frac{q_{\alpha}}{\tilde{q}_{\alpha}}-1\right\vert \leq\frac
{K}{c_{\alpha}}\cdot\frac{1}{c_{\alpha}\tilde{q}_{\alpha}}\text{,}%
\]
where $K$ is given in Proposition.\ref{prop-04}. Use Lemma.\ref{lem-e1},
\ref{lem-e2}, the result follows trivially.
\end{proof}

%

%TCIMACRO{\TeXButton{endshaded}{\end{shaded}}}%
%BeginExpansion
\end{shaded}%
%EndExpansion

\section{Experiment}

The idea presented in the previous section is illustrated through a simple
experiment. The environment is an MDP consisting of two groups of densely
connected states (cliques) linked by a long corridor. The agent has two
actions allowing it to move along the corridor deterministically, whereas the
transition probabilities inside each clique are randomly generated. The agent
assumes Dirichlet priors over all transition probabilities, and the goal is to
learn the transition model of the MDP. In the experiment, each clique consists
of $5$ states, (states $1$ to $5$ and states $56$ to $60$), and the corridor
is of length $50$ (states $6$ to $55$). The prior over each transition
probability is $Dir\left(  \frac{1}{60},\ldots,\frac{1}{60}\right)  $.

We compare four different algorithms: i) random exploration, where the agent
selects each of the two actions with equal probability at each time step; ii)
Q-learning with the immediate information gain $g\left(  ao\Vert h\right)  $
as the reward; iii) greedy exploration, where the agent chooses at each time
step the action maximizing $g\left(  a\Vert h\right)  $; and iv) a
dynamic-programming (DP) approximation of the optimal Bayesian exploration,
where at each time step the agent follows a policy which is computed using
policy iteration, assuming that the dynamics of the MDP is given by the
current posterior, and the reward is the expected information gain $g\left(
a\Vert h\right)  $.

\begin{figure}[h]
\begin{center}
\includegraphics[
width=4in]{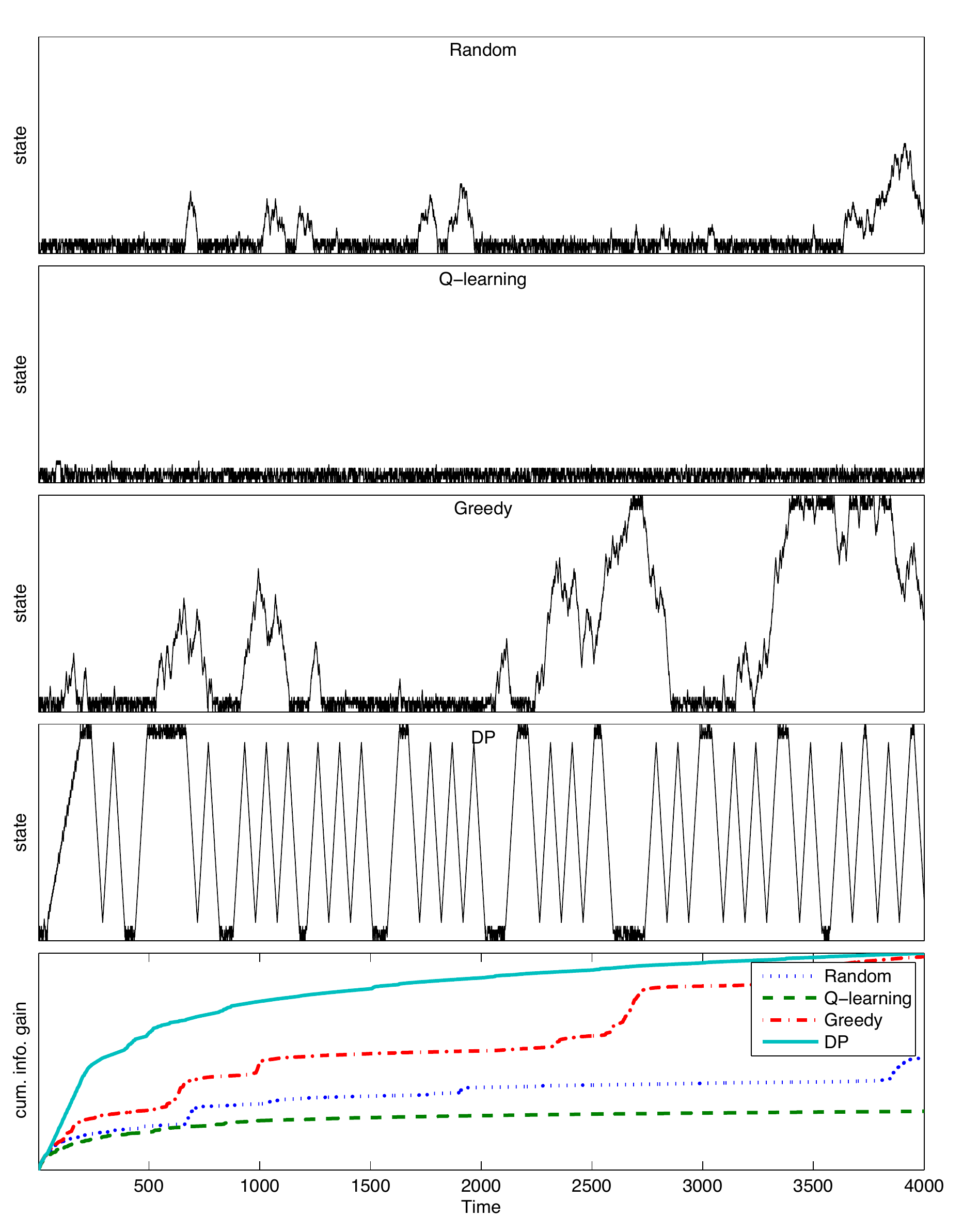}\vspace*{-5mm}
\end{center}
\caption{The exploration process of a typical run of $4000$ steps. The upper
four plots shows the position of the agent between state $1$ (the lowest) and
$60$ (the highest). The states at the top and the bottom correspond to the two
cliques, and the states in the middle correspond to the corridor. The lowest
plot is the cumulative information gain with respect to the prior.}%
\label{fig-exp}%
\end{figure}

Fig.\ref{fig-exp} shows the typical behavior of the four algorithms. The upper
four plots show how the agent moves in the MDP starting from one clique. Both
greedy exploration and DP approximation move back and forth between the two
cliques. Random exploration has difficulty moving between the two cliques due
to the random walk behavior in the corridor. Q-learning exploration, however,
gets stuck in the initial clique. The reason for is that since the jump on the
corridor is deterministic, the information gain decreases to virtually zero
after only several attempts, therefore the Q-value of jumping into the
corridor becomes much lower than the Q-value of jumping inside the clique. The
bottom plot shows how the cumulative information gain grows over time, and how
the DP approximation clearly outperforms the other algorithms, particularly in
the early phase of exploration.

\section{Related Work}

The idea of actively selecting queries to accelerate learning process has a
long history \cite{95-BayesExp,72-Fedorov,10-JuergenCuriosSurvey}, and
received a lot of attention in recent decades, primarily in the context of
active learning \cite{10-ALSettles} and artificial curiosity
\cite{91-SingaporeCur}. In particular, measuring learning progress using KL
divergence dates back to the 50's \cite{56-KLInfoGain,72-Fedorov}. In 1995
this was combined with reinforcement learning, with the goal of optimizing
future expected information gain \cite{95-RDIA}. Others renamed this Bayesian
surprise \cite{06-BayesSurpNIPS}.

Our work differs from most previous work in two main points: First, like in
\cite{95-RDIA}, we consider the problem of exploring a dynamic environment,
where actions change the environmental state, while most work on active
learning and Bayesian experiment design focuses on queries that do not affect
the environment \cite{10-ALSettles}. Second, our result is theoretically sound
and directly derived from first principles, in contrast to the more heuristic
application \cite{95-RDIA} of traditional reinforcement learning to the
problem of maximizing expected information gain. We formulated the concept of
curiosity (Q) value, and highlighted the necessity of balancing immediate
information gain and long-term expected information gain (see Eq.\ref{eq-04}).
In particular, we pointed out a previously neglected subtlety of using KL
divergence as learning progress.

Conceptually, however, our work is closely connected to artificial curiosity
and intrinsically motivated reinforcement learning
\cite{91-SingaporeCur,04-IntrinsRLNIPS,10-JuergenCuriosSurvey} for agents that
actively explore the environment without external reward signal. In fact, the
very definition of the curiosity (Q) value permits a firm connection between
pure exploration and reinforcement learning.

\section{Conclusion}

We have presented the principle of optimal Bayesian exploration in dynamic
environments, centered around the concept of the curiosity (Q) value. Our work
provides a theoretically sound foundation for designing more effective
exploration strategies. Based on this result, we establish the optimality of
the DP approximation of the optimal Bayesian exploration in the MDP case.%

\bibliographystyle{plain}
\bibliography{short}

\begin{thebibliography}{10}

\bibitem{97-IneqPsi}
Horst Alzer.
\newblock On some inequalities for the gamma and psi functions.
\newblock {\em Mathematics of Computation}, 66(217):373--389, 1997.

\bibitem{95-BayesExp}
Kathryn Chaloner and Isabella Verdinelli.
\newblock Bayesian experimental design: A review.
\newblock {\em Statistical Science}, 10:273--304, 1995.

\bibitem{72-Fedorov}
V.~V. Fedorov.
\newblock {\em Theory of optimal experiments}.
\newblock Academic Press, 1972.

\bibitem{06-BayesSurpNIPS}
L.~Itti and P.~F. Baldi.
\newblock Bayesian surprise attracts human attention.
\newblock In {\em NIPS'05}, pages 547--554, 2006.

\bibitem{56-KLInfoGain}
D.~V. Lindley.
\newblock On a measure of the information provided by an experiment.
\newblock {\em Annals of Mathematical Statistics}, 27(4):986--1005, 1956.

\bibitem{01-KLDir}
W.D. Penny.
\newblock Kullback-liebler divergences of normal, gamma, dirichlet and wishart
  densities.
\newblock Technical report, Wellcome Department of Cognitive Neurology,
  University College London, 2001.

\bibitem{91-SingaporeCur}
J\"urgen Schmidhuber.
\newblock Curious model-building control systems.
\newblock In {\em IJCNN'91}, volume~2, pages 1458--1463, 1991.

\bibitem{10-JuergenCuriosSurvey}
J\"urgen Schmidhuber.
\newblock Formal theory of creativity, fun, and intrinsic motivation
  (1990-2010).
\newblock {\em Autonomous Mental Development, IEEE Transactions on},
  2(3):230--247, 9 2010.

\bibitem{10-ALSettles}
Burr Settles.
\newblock Active learning literature survey.
\newblock Technical report, 2010.

\bibitem{04-IntrinsRLNIPS}
S.~Singh, Ag~Barto, and N.~Chentanez.
\newblock Intrinsically motivated reinforcement learning.
\newblock In {\em NIPS'04}, 2004.

\bibitem{95-RDIA}
Jan Storck, Sepp Hochreiter, and J\"urgen Schmidhuber.
\newblock Reinforcement driven information acquisition in non-deterministic
  environments.
\newblock In {\em ICANN'95}, 1995.

\end{thebibliography}

\end{document}